\documentclass{article}

\usepackage[table]{xcolor}
\usepackage[final]{corl_2026} %

\usepackage{hyperref}
\usepackage{enumitem}
\usepackage{amsmath}
\usepackage{amsfonts}
\usepackage{graphicx}
\usepackage{fontawesome5} %
\usepackage{marvosym} %
\usepackage{array}        %
\usepackage{tabularx}     %
\usepackage{wrapfig}
\usepackage{bbm}
\usepackage{duckuments}

\usepackage[colorinlistoftodos]{todonotes}
\definecolor{adoptEstablished}{RGB}{199,233,192} %
\definecolor{adoptRamping}{RGB}{253,212,158}     %
\definecolor{adoptFringe}{RGB}{217,217,217}      %
\newcolumntype{I}{>{\centering\arraybackslash}m{0.55cm}}
\newcolumntype{H}{>{\setbox0=\hbox\bgroup}c<{\egroup}@{}}  %

\newenvironment{myitemize}
{ \begin{itemize}[leftmargin=2em]
    \setlength{\itemsep}{3pt}
    \setlength{\parskip}{2pt}
    \setlength{\parsep}{3pt}     }
{ \end{itemize}                  } 

\newcommand{\squishlist}{
	\begin{list}{$\bullet$}
		{ \setlength{\itemsep}{0pt}
			\setlength{\parsep}{1pt}
			\setlength{\topsep}{1pt}
			\setlength{\partopsep}{0pt}
			\setlength{\leftmargin}{1.5em}
			\setlength{\labelwidth}{1em}
			\setlength{\labelsep}{0.5em} } }
\newcommand{\squishend}{\end{list} 
}

\newcommand{\tildeNice}{{\raise.17ex\hbox{$\scriptstyle\sim$}}}
\newcommand\lib{VisionSIM}

\title{ \vspace{-0.15in}\lib{}: Beyond Idealized Sensors — A Multi-Modal Simulator for Robot Perception}

\vspace{-0.75em}
\author{%
  Sacha Jungerman\textsuperscript{1}\quad
  Kaustubh Sadekar\textsuperscript{2}\quad
  Avery Gump\textsuperscript{1}\quad
  Sriram Narayanan\textsuperscript{3}\\[0.25em]
  \textbf{%
    Shantanu Gupta\textsuperscript{1}\quad
    Bhuyashi Deka\textsuperscript{1}\quad
    Srinivasa Narasimhan\textsuperscript{3}\quad
    Atul Ingle\textsuperscript{2}\quad
    Mohit Gupta\textsuperscript{1}%
  }\\[0.75em]
  \textsuperscript{1}University of Wisconsin -- Madison\quad
  \textsuperscript{2}Portland State University\quad
  \textsuperscript{3}Carnegie Mellon University\\
}

\begin{document}
\maketitle

\vspace{-2.15em}

\begin{abstract}
Robotic perception increasingly relies on heterogeneous sensor suites combining cameras with emerging modalities (event, SPAD, iToF/dToF, and thermal). Yet existing simulators idealize these sensors, hiding critical modality-specific failure modes that limit real-world deployment. We present \lib{}, a unified, physically grounded simulation platform that elevates sensor realism to a first-class concern by modeling the raw physical measurement process for diverse conventional and emerging modalities. We expose fine-grained control over physical sensor parameters, reproducing failure modes overlooked by general-purpose simulators: pileup and glare for LiDAR; phase wrapping, saturation, and motion artifacts for iToF; refractory drops for event cameras; and history-dependent emission for thermal cameras. This enables principled cross-modality comparison on identical scenes and trajectories. We release the platform and a large indoor dataset with synchronized raw measurements and rich ground-truth annotations.

\end{abstract}
\vspace{-1em}

\section{Introduction}
\label{sec:intro}

\vspace{-0.125in}
A robotics simulator presented with a scene will typically render a sharp, noise-free image. Yet, the same scene captured by a real conventional camera might show heavy motion blur from a long exposure; a real single-photon LiDAR aimed at a retroreflective object might report a phantom obstacle in front of it; a long-wave infrared (LWIR) thermal camera will show a lingering thermal footprint of a warm handprint on a desk or a radiator still glowing minutes after shutoff. These are not exotic edge cases --- they are the operating conditions that decide whether a deployed robot succeeds or fails. However, they are largely absent from today's general-purpose robotics simulators.

Modern simulation platforms have transformed how robots are trained, evaluated, and deployed: physics engines such as MuJoCo~\cite{todorov2012mujoco}, Isaac Gym~\cite{issacgym}, and Genesis~\cite{Genesis} model rigid-body dynamics with impressive fidelity, while rendering engines such as Unreal~\cite{unrealengine} and Blender~\cite{blender} bring photorealistic digital twins of factories and cities within reach. However, these platforms share a critical blind spot: they do not faithfully simulate how agents \emph{actually see the world}. Real-world agents act on what their sensors deliver, not on ground truth, and typically combine multiple complementary sensors (Table~\ref{tab:comp-imaging-final-icons}). Yet most simulation engines do not support the necessary breadth of sensors, or they model them only superficially: ideal cameras in a well-lit environment, LiDARs as noiseless range-finders, IMUs as perfect integrators, thermal cameras as static temperature textures. Such simplifications ignore the noise, motion blur, and modality-specific artifacts that dominate real measurements --- from glare and pileup in single-photon LiDAR~\cite{Gupta_2019b} to the geometry-, material-, and history-dependent emission that defines real thermal imagery.
 
Existing efforts address this gap only partially. Sensor-specific simulators, such as ESIM~\cite{esim} and v2e~\cite{Hu2021-v2e} for event cameras, are highly specialized and cannot support cross-modality comparisons. Physically-based infrared generators such as MuSES~\cite{upadhyay2024comprehensivesurveysyntheticinfrared} and DIRSIG~\cite{schott1999dirsig} couple heat-transfer solvers to radiometric rendering, but target terrain-scale signature prediction driven by weather and atmospheric models, rather than the object-scale thermal phenomena central to embodied perception. General-purpose platforms such as CARLA~\cite{carla}, AirSim~\cite{airsim}, Isaac Gym~\cite{issacgym}, and Habitat~\cite{habitat} expand sensor coverage but inherit the same idealizations. As a result, robot system architects cannot compare candidate sensor suites on common ground without expensive hardware prototyping.

\begin{table*}[t!]
\scriptsize
\begin{tabularx}{\textwidth}{
  |c l l l X X c|
}
\hline
\multicolumn{2}{|c}{\textbf{Modality}} &
\multicolumn{1}{c}{\textbf{Raw Measurement}} &
\multicolumn{1}{c}{\textbf{Recoverable Scene Properties}} &
\textbf{Sim.} &
\textbf{HW} &
\textbf{Outlook} \\
\hline

\rowcolor{adoptEstablished}
\faIcon{camera} & RGB cameras* &
Visible radiance &
Appearance, texture, coarse geometry cues &
\faIcon{star} &
\faIcon{star} &
{\tiny \faIcon{stop}} \\

\rowcolor{adoptEstablished}
\faIcon{images} & Stereo / multi-view* &
Multi-view radiance, parallax &
Metric 3D geometry, surface layout &
\faIcon{star} &
\faIcon{star} &
{\tiny \faIcon{stop}} \\

\rowcolor{adoptRamping}
\Laserbeam & LiDAR (SPAD)* &
Photon arrival times; transients &
Low-light depth, long-range geometry &
\faIcon{star-half-alt} &
\faIcon{star-half-alt} &
\faIcon{forward} \\

\rowcolor{adoptRamping}
\faIcon{wave-square} & LiDAR (iToF)* &
Correlation measurements &
Depth, dynamic structure &
\faIcon[regular]{star-half-alt} &
\faIcon[regular]{star} &
\faIcon{forward} \\

\rowcolor{adoptRamping}
\faIcon{satellite-dish} & Radar (mmWave) &
RF phase, Doppler, reflections &
Coarse geometry, velocity, micro-motion &
\faIcon[regular]{star} &
\faIcon{star} &
{\tiny \faIcon{play}} \\

\rowcolor{adoptRamping}
\faIcon{thermometer-half} & Thermal (LWIR)* &
Thermal emission (8--14 $\mu$m) &
Temperature fields, heat signatures &
\faIcon{star} &
\faIcon{star} &
{\tiny \faIcon{play}} \\

\rowcolor{adoptRamping}
\faIcon{clock} & Event cameras* &
Asynchronous intensity changes &
High-speed motion, edges, dynamics &
\faIcon{star-half-alt} &
\faIcon[regular]{star} &
\faIcon{forward} \\

\rowcolor{adoptRamping}
\faIcon{hand-paper} & Inertial Unit* &
Forces, acceleration, joint states &
Contact state, pose, inertial dynamics &
\faIcon{star} &
\faIcon{star} &
{\tiny \faIcon{play}} \\

\rowcolor{adoptFringe}
\faIcon{volume-up} & Audio / acoustic &
Pressure waves and echoes &
Occluded structure, material \& cavity cues &
\faIcon[regular]{star} &
\faIcon[regular]{star} &
{\tiny \faIcon{play}} \\

\rowcolor{adoptFringe}
\faIcon{palette} & Hyperspectral &
Spectral reflectance signatures &
Material, chemical, biological composition &
\faIcon[regular]{star} &
\faIcon[regular]{star} &
\faIcon{forward} \\

\hline
\end{tabularx}

\vspace{0.2em}
\scriptsize
\textbf{Adoption:}
\colorbox{adoptEstablished}{\strut Established} \,
\colorbox{adoptRamping}{\strut Ramping Up} \,
\colorbox{adoptFringe}{\strut Fringe} \quad
\textbf{Maturity:}
\faIcon[regular]{star}~Low \,
\faIcon{star-half-alt}~Med \,
\faIcon{star} High \quad
\textbf{Outlook:}
{\tiny \faIcon{stop}}~stable \,
{\tiny \faIcon{play}}~growth \,
\faIcon{forward}~accelerating

\vspace{-0.75em}
\caption{\small{{\bf Computational imaging modality taxonomy.} Each modality is characterized by its \emph{raw physical measurement} and the \emph{recoverable scene properties}. The Sim. and HW columns indicate simulation and hardware maturity, respectively. Outlook glyphs indicate likely near- to mid-term trajectories in availability and adoption. *Currently supported by \lib{}.}}
\label{tab:comp-imaging-final-icons}
\vspace{-0.2in}
\end{table*}

\begin{figure*}[t]
    \centering
    \includegraphics[width=1.00\textwidth]{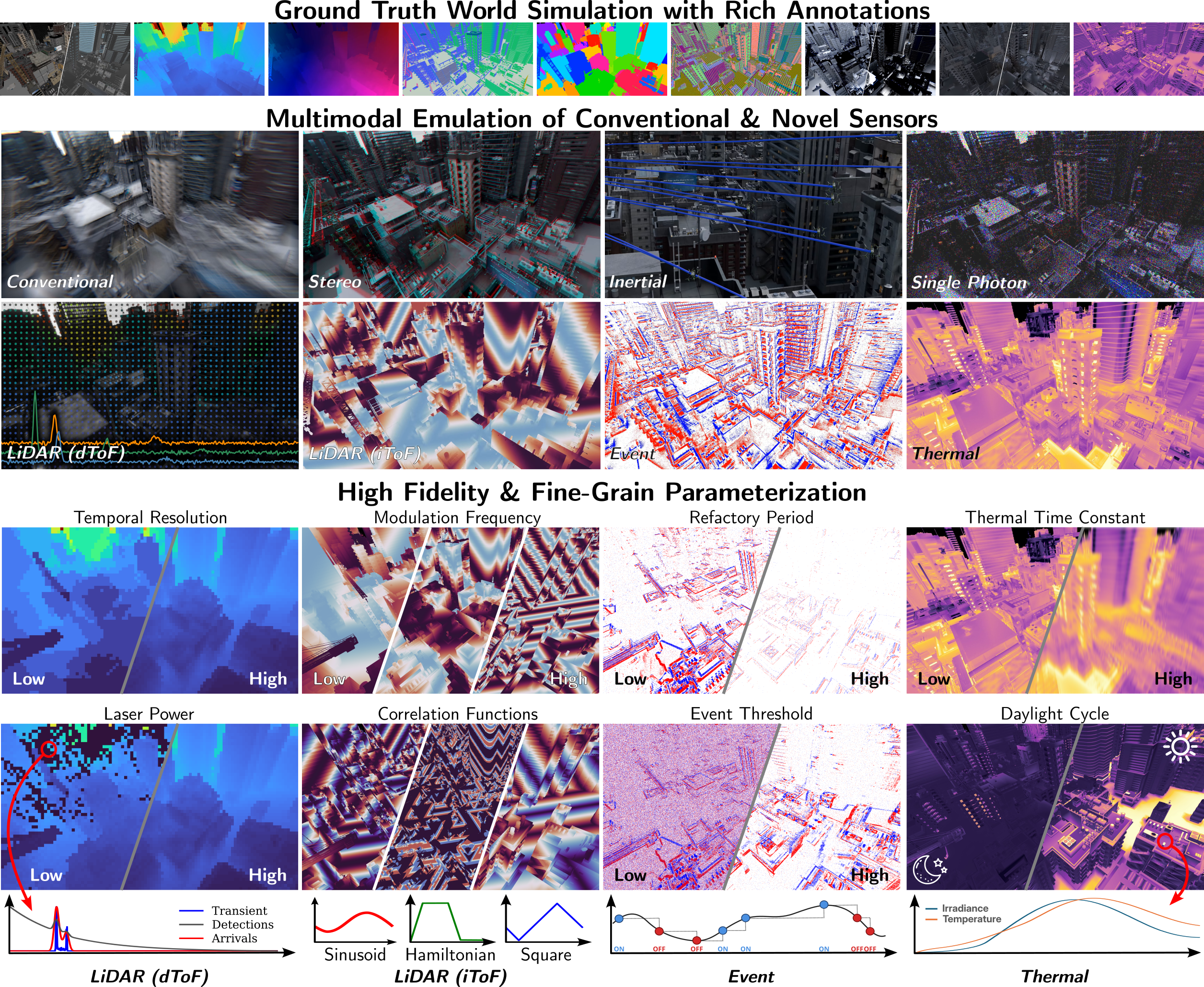}
    \vspace{-1.5em}
    \caption{\small{\textbf{Multi-modal, high-fidelity and controllable sensor emulation.} Robotic simulation pipelines often focus on scene content, overlooking physically accurate sensor modeling. \lib{} aims to bridge this gap by emulating diverse sensor types and their failure modes. (Top) A single scene is rendered once with dense ground truth — color, depth, optical flow, surface normals, segmentation, point map, direct and indirect illumination, and a temperature map. (Middle) The same scene is then replayed through a breadth of sensor models. (Bottom) Each model exposes and enables sweeping across the sensor's physical parameters (e.g., thermal time constant, correlation function, event thresholds), thereby reproducing modality-specific failure modes.   \label{fig:teaser}
    \vspace{-1.5em}
    }}
\end{figure*}

Our guiding hypothesis is that a high-fidelity simulator must model not just world physics and geometry, but also perception itself --- with realistic sensors that bridge photons to measurement. To this end, we present \lib{}, a unified, physically grounded sensor simulation platform for robotics that aims to elevate sensor realism to a first-class concern, on par with scene and physics modeling (Figure~\ref{fig:teaser}). \lib{} is built around the following key design goals:
 
\textbf{Breadth across conventional and emerging sensors.} \lib{} simulates conventional and high-speed cameras, stereo rigs, passive single-photon avalanche diode (SPAD) cameras, direct and indirect time-of-flight (dToF/iToF) LiDARs, event cameras, long-wave infrared (LWIR) thermal cameras, and inertial measurement units (Table~\ref{tab:comp-imaging-final-icons}). Candidate sensor suites can be evaluated on identical scenes and trajectories, enabling principled comparisons across a wide range of sensors.
 
\textbf{High physical fidelity, including failure modes.} \lib{} emulates raw measurements for each modality (e.g., transient histograms, correlation taps, and asynchronous brightness changes). This naturally reproduces key failure modes: SPAD pileup and glare; iToF wrapping and motion artifacts; event refractory limits; and path-tracing-coupled dynamic thermal heat transport, capturing failure modes that, to our knowledge, lie beyond the reach of existing general-purpose simulators.
 
\textbf{Fine-grained control over sensing parameters.} \lib{} exposes the physical parameters that govern each modality's measurement process (Table~\ref{tab:sensor_models}): iToF correlation functions; event contrast thresholds; dead time and quantum efficiency for SPADs; microbolometer thermal time constants, and per-surface emissivity for thermal cameras; exposure, gain, and ISP parameters for conventional cameras. This parameterization lets a user not only sweep across sensor designs, but also match the simulator to a specific physical device --- the basis for calibrated \emph{perceptual} digital twins.

We demonstrate \lib{} by constructing a digital twin of a real single-photon LiDAR, recovering some of its key parameters directly from captured data. This reduces the mean sim-to-real depth error and reproduces two failure modes not previously simulated at this fidelity: pileup-induced range walk, and point-cloud ghosting from retroreflective glare. We further show how \lib{} can be used to augment existing datasets and to evaluate competing sensor suites (Sec.~\ref{sec:applications}). These case studies focus on open-loop sensor fidelity; closed-loop validation is discussed in Section~\ref{sec:outlook}. Finally, we release an indoor dataset featuring synchronized raw outputs of all supported sensors and ground truth annotations (Table~\ref{tab:dataset_comparison}). To our knowledge, it is among the most comprehensive multi-modal datasets of its kind, supporting both training and benchmarking of sensor-fusion algorithms.

\vspace{-1.0em} 
\section{Related Works}
\label{sec:related-work}
\vspace{-1.0em}

\noindent \textbf{Task Specific Datasets \& Pipelines:} There exist several publicly available real and synthetic datasets supporting various computer vision tasks. These are often specific to applications, such as autonomous driving~\cite{nuscenes2019,a2d2,waymo,kitti}, or tasks, such as optical flow or stereo matching~\cite{sintel,spring,Middlebury2014,flyingthings}. They often lack the metadata, ground truth, and raw sensor access needed for use beyond their original task. The ones with rich annotations are either subsets of larger datasets~\cite{a2d2,waymo}, or are synthetically generated using purpose-built engines such as CARLA~\cite{carla}, which are not easily generalizable to other tasks and may lack unconventional modalities such as single-photon or thermal sensors. 
 
\noindent \textbf{General Purpose Simulation Engines:} Many engines (Isaac Gym, Genesis, PhysX, Kubric, Bullet)~\cite{issacgym,Genesis,physx2020,kubric,bullet} prioritize fast and accurate physics for robotics learning, often at the cost of rendering fidelity. Frameworks such as BlendProc~\cite{blendproc} and Kubric~\cite{kubric} yield photorealistic results, yet perform no sensor emulation. Others, such as ESIM~\cite{esim}, target a single sensor type. \lib{} addresses both aspects --- sensor breadth with physical realism --- within a unified pipeline. 
 
\noindent \textbf{Sensor Emulation Methods:} Emulating different sensor types or characteristics from existing real data is an active research direction, lowering the barrier to adopting non-standard sensors. v2e~\cite{Hu2021-v2e} converts videos to event streams; physics-based infrared generators such as DIRSIG~\cite{schott1999dirsig} and MuSES~\cite{upadhyay2024comprehensivesurveysyntheticinfrared} couple heat solvers to rendering, but are proprietary or domain-specific. \lib{} combines these trends by bringing single-photon~\cite{bruschini2019singlephoton,qbp,Ingle_CVPR2019_highflux}, transient, and light-driven thermal~\cite{maniCVPR24,TanakaFar} sensing into a unified, multi-modal simulation framework for embodied perception.

\vspace{-0.5em} 
\section{Architecture Overview}
\label{sec:method_overview}

\vspace{-0.1in}
\lib{}'s key design choice is to \emph{decouple} photorealistic rendering from sensor emulation. We first render the clean scene, and then apply physical non-idealities to produce raw measurements. Because these layers are separate, one rendered sequence can be replayed through any sensor model, and parameters can be tweaked at zero extra rendering cost, enabling rapid device calibration.

\textbf{World Simulation Layer.} This layer is the image-formation engine, producing photorealistic reference frames together with dense ground-truth annotations (depth, normals, optical flow, segmentation, point maps, illumination, poses, and intrinsics) from a Blender scene. It additionally generates ground-truth temperature fields through an integrated, GPU-accelerated heat-transport solver that drives the thermal sensor model (Appendix~\ref{sec:thermal-rendering}). A central challenge is producing frames at the extreme rates demanded by high-speed and asynchronous sensors — event and single-photon cameras operate at tens to hundreds of thousands of frames per second — which is prohibitively expensive to render. \lib{} addresses this with distributed client/server rendering across heterogeneous hardware, a \emph{keyframe-remapping} technique that temporally dilates animation splines so complex physics can be sampled at arbitrary frame rates, and a render-then-interpolate strategy that renders ground-truth frames only at each scene's intrinsic \emph{native frame rate} and synthesizes the rest with deep video interpolation (e.g., RIFE~\cite{huang2022rife}). Together, these yield order-of-magnitude reductions in wall-clock time while preserving high fidelity to fully rendered sequences (Appendix~\ref{app:world-sim}-\ref{app:interpolation}).

\textbf{Sensor Emulation Layer.} This layer ingests the noise-free, high-frame-rate data and transforms them into realistic sensor measurements. Rather than applying generic additive Gaussian noise to idealized outputs, it implements physically constrained, per-modality measurement models from which noise, distortions, and failure modes naturally emerge as physical consequences. The layer supports conventional and high-speed cameras, stereo and multi-camera rigs, passive single-photon cameras, single-photon (dToF) LiDAR, indirect time-of-flight (iToF) LiDAR, event cameras, thermal cameras, and inertial measurement units. Since it operates on shared reference frames, any subset can be co-simulated on identical scenes and trajectories. The resulting multi-modal measurements are saved in a compressed, sensor-aware format compatible with downstream learning frameworks~\cite{nerfstudio,brush}; stochastic binary single-photon frames, in particular, are stored in a bit-packed manner that achieves $8\times$ lossless compression with on-the-fly random access (Appendix~\ref{app:datastorage}).

\section{Physically-Grounded Sensor Models and Failure Modes}
\label{sec:sensor_emulation}

\vspace{-0.1in}
We now detail the physically accurate sensor models for three representative modalities whose failure modes are least faithfully captured by existing simulators: single-photon LiDAR, indirect time-of-flight LiDAR, and thermal cameras. Forward models and implementation details for all supported sensors are in Appendix~\ref{app:sensors}; Table~\ref{tab:sensor_models} summarizes their failure modes and parameters.
 
\begin{table*}[t!]
\centering
\fontsize{7.65pt}{6.5pt}\selectfont
\setlength{\tabcolsep}{1.5pt}
\begin{tabularx}{\textwidth}{
  | l 
    >{\raggedright\arraybackslash\hsize=1.03\hsize}X   %
    >{\raggedright\arraybackslash\hsize=0.97\hsize}X | %
}
\hline
\rule{0pt}{5.9pt}
\textbf{Modality} &
\textbf{Modeled Non-idealities \& Failure Modes} &
\textbf{Key Exposed Parameters} \\
\hline
\faIcon{camera} ~ RGB cameras &
Shot/read noise, motion blur, \newline full-well saturation, quantization &
Exposure, gain, full-well capacity, \newline bit depth, ISP artifacts\\
\faIcon{video} ~ High-speed cameras &
Elevated read noise, low signal, motion blur &
Exposure, gain, read noise \\
\faIcon{images} ~ Stereo / multi-view &
Disparity error, inherited camera non-idealities &
Baseline, intrinsics, per-camera parameters \\
\Laserbeam ~ LiDAR (SPAD, dToF) &
Pileup, range walk, count saturation, \newline ambient background, glare &
IRF, dead time, quantum efficiency, \newline ambient rate, TDC resolution \\
\faIcon{wave-square} ~ LiDAR (iToF) &
Phase wrapping, saturation, ambient corruption, multi-tap \& intra-tap motion, noise, quantization &
Coding function, modulation frequency, \newline taps $K$, full-well capacity, read noise, ADC bits \\
\faIcon{lightbulb} ~ Passive single-photon &
Shot-noise, quantum efficiency, dark counts &
Exposure, dark count rate, fill factor \\
\faIcon{thermometer-half} ~ Thermal (LWIR) &
Temperature-dependent emission, thermal inertia &
Thermal parameters, FEM simulation \\
\faIcon{clock} ~ Event cameras &
threshold variation, refractory drop-out, shot noise &
Contrast threshold, refractory period \\
\faIcon{hand-paper} ~ IMU &
Random-walk drift, bias instability, Gaussian noise &
Noise, bias instability, output rate \\
\hline
\end{tabularx}
\vspace{-0.05in}
\caption{\small{{\bf Failure modes and parameters modeled by \lib{}.} For each supported modality, we list the non-idealities and failure modes \lib{} reproduces and the principal user-controllable parameters. Full forward models for all modalities are given in Appendix~\ref{app:sensors}.}}
\label{tab:sensor_models}
\vspace{-0.2in}
\end{table*}

\vspace{-0.05in}
\subsection{Single-Photon (dToF) LiDAR} \label{subsec:spad} \vspace{-0.1in}

Commercial solid-state LiDAR is rapidly adopting single-photon avalanche diode (SPAD) arrays~\cite{ouster,hesai} to capture photon-arrival histograms for depth estimation. \lib{} physically models the histogram acquisition pipeline. For each pixel, the discrete \emph{scene response} $T[b]$ at bin index $b$ aggregates the radiometric contributions of scene points within a field-of-view mask $\mathcal{M}$ as
\begin{equation}
    T[b] = \eta_\text{opt} \sum_{p \in \mathcal{M}} \mathbb{I}\big(b \Delta t \le t_p < (b+1)\Delta t\big)\, \frac{\alpha_p}{z_p^2}\, W(p),
    \label{eq:transient}
\end{equation}
where $\Delta t$ is the temporal bin width, $\alpha_p$ is albedo, $1/z_p^2$ the inverse-square falloff, $W(p)$ the mask weight, and $\eta_\text{opt}$ the quantum efficiency. The spatially-resolved mask $\mathcal{M}$ models common optical effects like vignetting and \emph{blooming} from retroreflective returns (Section~\ref{sec:applications}). SPAD crosstalk is not yet modeled; it can be added as a per-pixel spatial kernel applied to the transient, which would capture optical and electrical coupling between neighbouring detectors. The total photon-arrival rate is then modeled as $\lambda(t) = (T * g)(t) + \lambda_\text{ambient}$, where $\lambda_\text{ambient}$ is a constant background rate, and $g(t)$ is the system's impulse response function (IRF), a calibrated or parametric laser pulse. Finally, incident photon counts are drawn from an inhomogeneous Poisson process governed by $\lambda(t)$.

In high-flux conditions, modeling \emph{dead time} (time interval after a photon detection when the sensor is inhibited) emulates pileup-distorted measurements~\cite{coates1968,oconnor-phillips}: later-arriving photons are systematically lost, resulting in histograms biased toward earlier bins and shallower depth estimates (\emph{range walk error}) and poor SNR, particularly for distant objects (Fig.~\ref{fig:teaser}). Intra-histogram motion blur can cause erroneous depth estimates, causing safety-critical issues (Fig.~\ref{fig:applications}). Conventional simulators that model LiDARs as noiseless range-finders would fail to capture these effects. 
 
\vspace{-0.05in}
\subsection{Indirect Time-of-Flight (iToF) LiDAR}
\label{subsec:itof} \vspace{-0.1in}
 
\begin{figure*}[t]
\begin{center}
\includegraphics[width=1.0\textwidth]{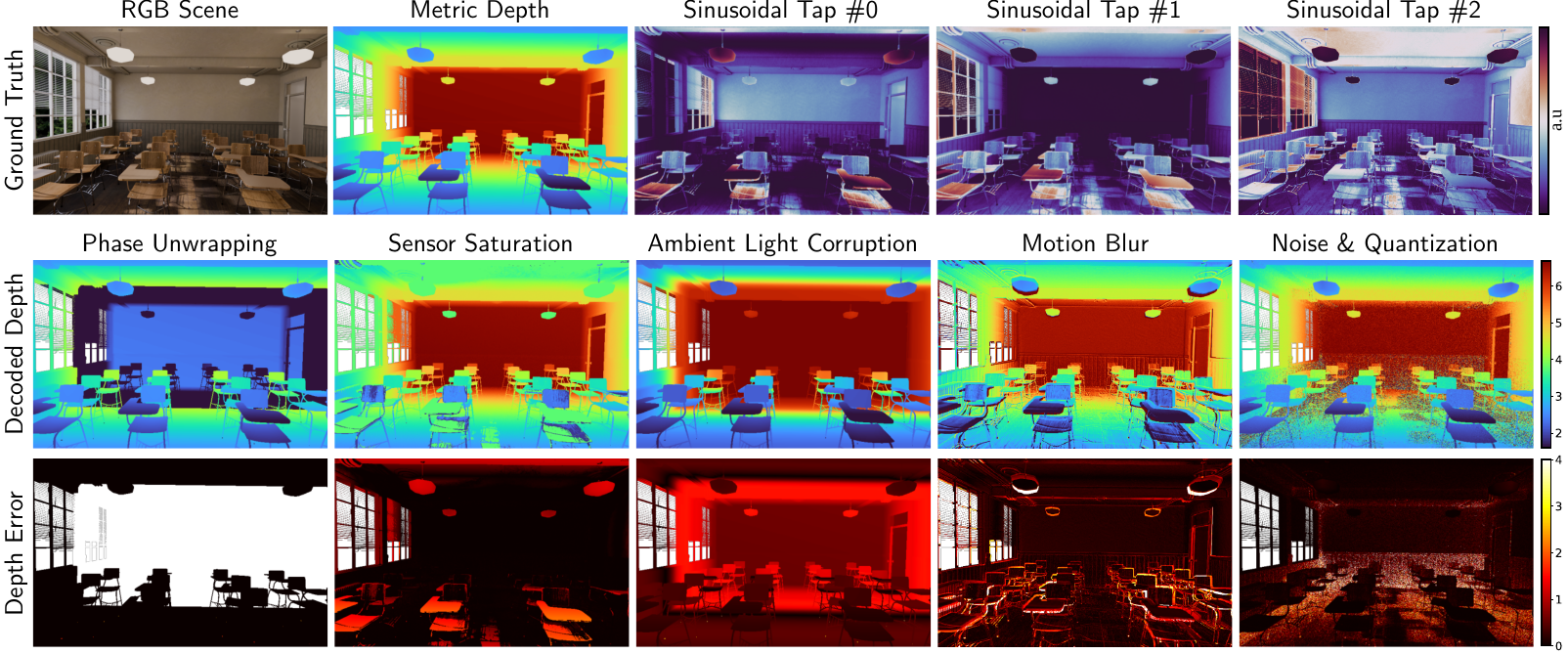}
\end{center}
\vspace{-1em}
\caption{\small{\textbf{Indirect Time-of-Flight (iToF) simulation and failure cases:} (\textbf{Top}) Ground truth RGB scene, ground truth depth map, and ideal correlation measurements (taps 1, 2, and 3) for a conventional sinusoidal coding scheme. (\textbf{Bottom}) Common iToF failure cases and their effects on depth reconstruction, including: incorrect modulation frequency (causing phase wrapping), sensor saturation (clipping correlation measurements), ambient light corruption (reducing signal-to-noise ratio, especially for farther objects), motion blur (multi-tap artifacts from scene dynamics), and sensor noise and quantization artifacts (high frequency depth errors).}}
\label{fig:itof}
\vspace{-0.2in}
\end{figure*}
 
Indirect, or continuous-wave, time-of-flight cameras are similarly widespread, shipping in consumer and industrial depth sensors from major manufacturers~\cite{sony,stmicro}. Rather than timing individual photons, they recover depth from the phase of amplitude-modulated light~\cite{lange2001}: the scene is illuminated with a periodically modulated source, and the sensor correlates the returning light against $K$ phase-shifted demodulation functions, producing $K$ \emph{correlation taps} from which depth is decoded. Following the continuous-wave image-formation model~\cite{GuptaToG_2018}, the $i$-th measured tap at a pixel is:
\begin{equation}
\tilde{B}_i = \mathcal{Q}\!\left(\beta\, F_i(\Gamma) + A + \eta_i\right), \quad i = 1,\dots,K,
\label{eq:itof}
\end{equation}
where $\Gamma$ is the scene depth, $F_i$ the normalized correlation function for tap $i$ (set by the coding scheme), $\beta$ includes albedo and distance-squared falloff, $A$ is the ambient component, $\eta_i$ is shot and read noise, and $\mathcal{Q}(\cdot)$ models sensing non-idealities (e.g., ADC quantization, saturation).
 
Equation~\ref{eq:itof} captures physical failure modes explicitly (Figure~\ref{fig:itof}): (i) \emph{Phase wrapping:} depth wraps beyond the unambiguous range, $\Gamma_\text{max} = c/2f$, set by the modulation frequency $f$, rendering distant objects close; (ii) \emph{Saturation:} measurements clip when signal exceeds full-well capacity; (iii) \emph{Ambient corruption:} large $A$ reduces modulation contrast, corrupting depth particularly for distant, dimly-lit points; (iv) \emph{Motion artifacts:} intra- and inter-tap motion (for sequential taps) cause spatial tearing and boundary ghosting, which we model by integrating temporally offset interpolated frames; (v) \emph{Noise \& quantization:} $\eta_i$ and $\mathcal{Q}$ cause high-frequency depth errors and banding.
 
\vspace{-0.05in}
\subsection{Thermal Cameras} \label{subsec:thermal} \vspace{-0.1in}
  
Long-wave infrared (LWIR) thermal cameras differ from other modalities in that they not only measure reflected thermal light due to light transport but also capture light emitted by an object due to the underlying heat transport process happening within and around the object. The radiance at each pixel is the integrated long-wave emission of the surface in front of it, governed by Planck's law and weighted by the surface's emissivity, with an additive reflected-ambient term. 
\begin{wrapfigure}{l}{0.4825\textwidth}
    \includegraphics[width=1.0\linewidth, trim={0 42cm 13cm 0}, clip]{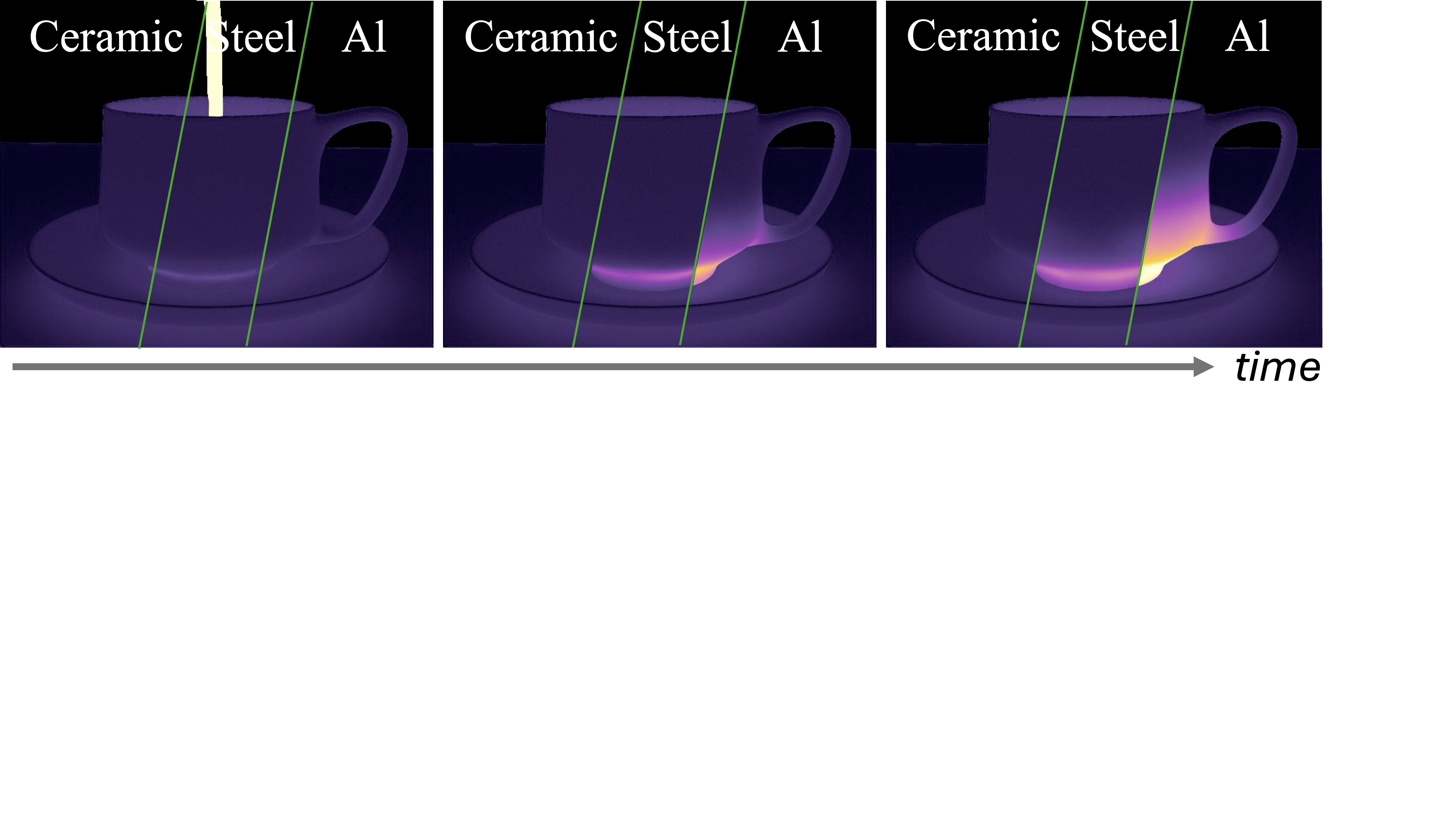}
    \vspace{-1.7em}
    \caption{\small{\textbf{Transient Thermal Simulation:} Thermal video of ceramic, steel, and aluminum cups over the 4 seconds following a hot-tea fill. The distinct heating rates reflect each material's thermal diffusivity ($\alpha = 2,\ 18.8,\ 97\ \mathrm{mm^2/s}$, respectively).}}
    \label{fig:thermal}
    \vspace{-0.85em}
    \label{fig:wrapfig}
\end{wrapfigure}
Crucially, this signal depends not on instantaneous lighting but on the temperature field that geometry, materials, and the scene's recent history (minutes to hours of prior illumination, contact, and conduction) have produced --- a quantity conventional graphics pipelines do not compute. Existing thermal-simulation tools~\cite{airsim} sidestep this by painting a static ``temperature texture'' onto each surface or assuming radiative equilibrium with a single user-provided heat source~\cite{Zombie}; neither approach reproduces the material- and history-dependent thermal contrast that defines real LWIR imagery, such as differential cooling across materials (ceramic vs. metal mugs cooling at different rates in Fig.~\ref{fig:thermal}), residual heat from recent contact, or radiator-like sources that linger after shutoff. \lib{} directly integrates a fast physics-based heat solver in its world simulation layer, so thermal imagery can be produced alongside the other modalities. For details on our light-to-heat simulation, thermal rendering, and microbolometer inertia, see Appendix~\ref{sec:thermal-rendering}.

\section{Case Studies: Design and Analysis of Robot Perception Systems}
\label{sec:applications}
\vspace{-1em}

\label{subsec:twin}

\begin{figure*}[t]
\begin{center}
\includegraphics[width=1.0\textwidth]{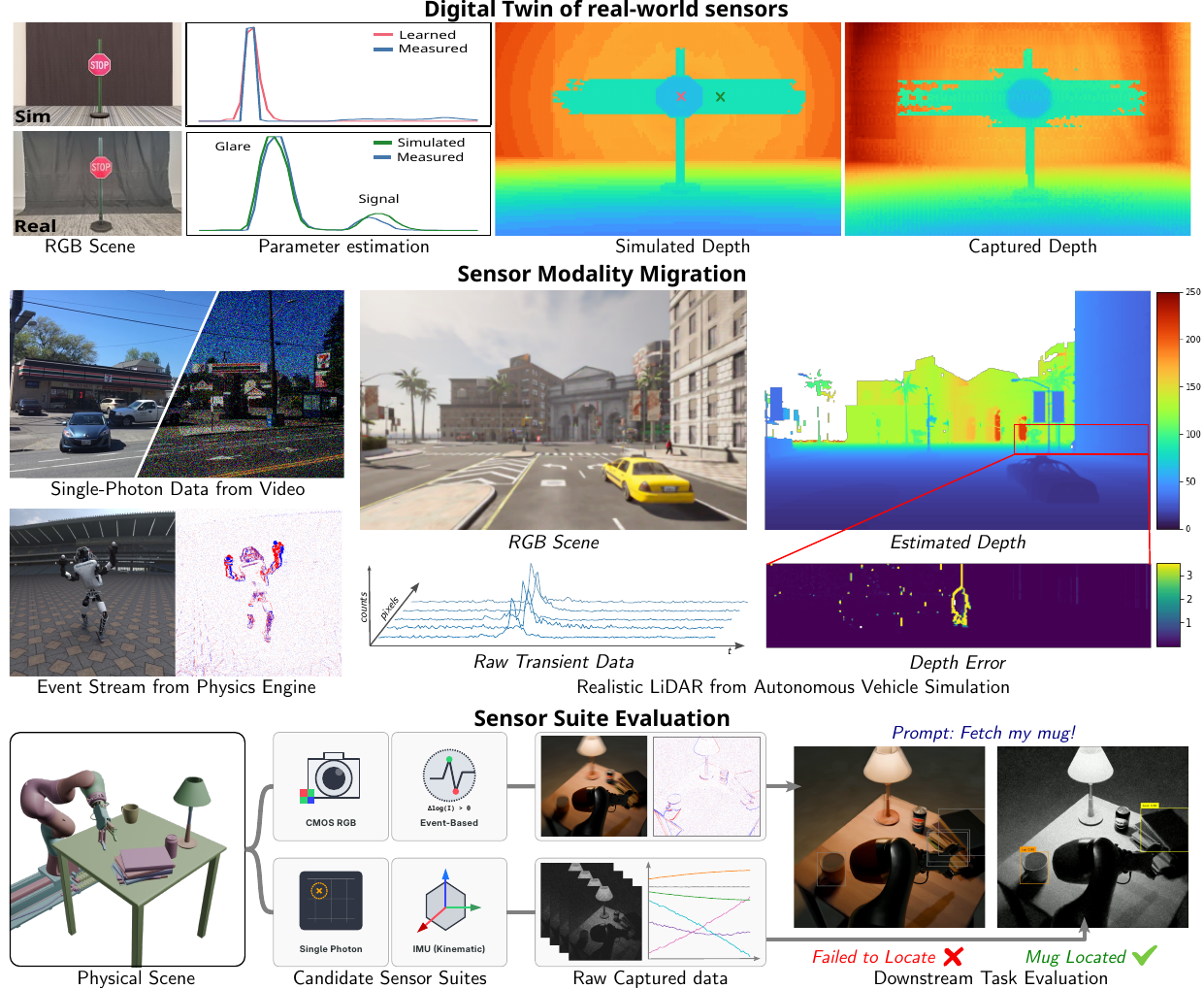}
\end{center}
\vspace{-1em}
\caption{\small{\textbf{Application Case-Studies:} (\textbf{Top}) A digital twin of a commercial LiDAR, where sensor parameters such as dead time and impulse response function are estimated from real-world data to correct for artifacts like range-walk. (\textbf{Middle}) Sensor modality migrations: \lib{} can uplift existing datasets (e.g., XVFI~\protect\cite{sim2021xvfi}) or high-speed renders to emulate new sensor modalities like single-photon or event cameras. (\textbf{Bottom}) Sensor suite evaluation: \lib{} enables rapid evaluation of different sensor configurations for perception tasks.}}
\label{fig:applications}
\vspace{-0.2in}
\end{figure*}

Next, we illustrate potential applications of \lib{}, each with a case study (Figure~\ref{fig:applications}) designed to serve as a proof-of-concept via a representative example. All studies are open-loop, focusing on sensor fidelity rather than closed-loop control (Section~\ref{sec:outlook}).

\noindent\textbf{Building digital twins of real sensors.} \lib{} enables building \emph{perceptual} digital twins: a model whose parameters are calibrated so that its simulated raw measurements match those of a specific physical device, capturing the sensor's characteristic noise and failure modes. This is critical for closing the sim-to-real gap on the target deployment hardware.
 
We build a digital twin of a commercial single-photon LiDAR (Adaps Photonics ADS6311~\cite{adaps_ads6311}; details in Appendix ~\ref{app:digital_Twin}). Because \lib{}'s dToF model is differentiable, we recover the device's parameters directly from real captures via gradient-based fitting of a measurement-matching loss, which we demonstrate with the sensor's impulse response function (IRF). We validate the twin on a scene captured in the lab and modeled in Blender (Figure~\ref{fig:applications}, top), chosen because it exercises two failure modes at once. The retroreflective sign produces an intense return that saturates the detector; the recovered IRF and dead time reproduce the resulting \emph{pileup} and range-walk error. That same intense return induces \emph{glare} through internal multipath between the sensor optics and the detector array, which the spatially-varying PSF captures, reproducing the characteristic point-cloud ghosting. The resulting reconstructed depth closely tracks the device's measured depth. Faithfully modeling the IRF, pileup, and glare lowers the RMSE of the digital twin by $\tildeNice1.53$ meters.
 
\label{subsec:migration}
\noindent\textbf{Migrating existing datasets to new sensing modalities.} The decoupled architecture also enables \emph{migrating} existing datasets to modalities they were never captured with, by bypassing world simulation and feeding pre-existing data directly to the sensor emulation layer. We show three examples (Figure~\ref{fig:applications}, middle): emulating a passive single-photon camera from the XVFI high-speed video dataset~\cite{sim2021xvfi}; generating an event stream from a Genesis render of a dancing humanoid~\cite{Genesis,wang2023robogen}; and producing raw single-photon LiDAR transients from a CARLA driving sequence~\cite{carla}. Uplifting a conventional autonomous-driving dataset into time-resolved single-photon measurements lets sensor-specific failure modes be exposed: here, the pedestrian might not be detected \emph{at the sensor level} as their lateral motion induces motion blur within the transient histograms, causing depth errors in excess of $3$ meters. This lets researchers obtain data for emerging sensors by reusing existing assets, rather than recapturing scenes with specialized and often unavailable hardware. Dataset migration is bounded by what the source data contains: the frame rate caps the temporal resolution of the emulated sensor, and each modality needs its own annotations, so a dataset without depth cannot be uplifted to iToF or dToF.

\label{subsec:suite}
\noindent\textbf{Evaluating \& benchmarking sensor suites.} Sensor emulation decoupling enables a single rendered scene to be replayed through many sensor configurations at no additional rendering cost, allowing comparisons of candidate suites before any hardware is built. We demonstrate this through a robot perception case study, where a system architect selects between two sensor suites for low-light, dynamic environments: a conventional camera paired with an event camera, or a passive single-photon camera and IMU. Both are emulated on identical high-speed trajectories; each suite's raw measurements are reconstructed into image-like representations using existing methods~\cite{Sun_2024_ECCV,qbp}, and passed to an off-the-shelf detection and tracking model~\cite{rf-detr} to compare downstream task performance (Figure~\ref{fig:applications}, bottom). Detection fails in the first case, while the second sensor suite yields $90\%$ confidence. Both reconstruction models and the detector are used zero-shot, so the result characterizes the full pipeline rather than the sensor suite alone; retraining on raw measurements would remove the reconstruction bias. Sweeping sensor parameters (exposure, event threshold, fill factor, etc.) within the same framework lets the architect map the trade-off across operating regimes, a comparison that is prohibitively expensive to perform with real hardware.

\section{Sim-to-Real Validation}
\label{sec:validation}
\vspace{-1em}

\begin{wrapfigure}{r}{0.55\textwidth}
    \vspace{-1.5em}
    \includegraphics[width=1.0\linewidth]{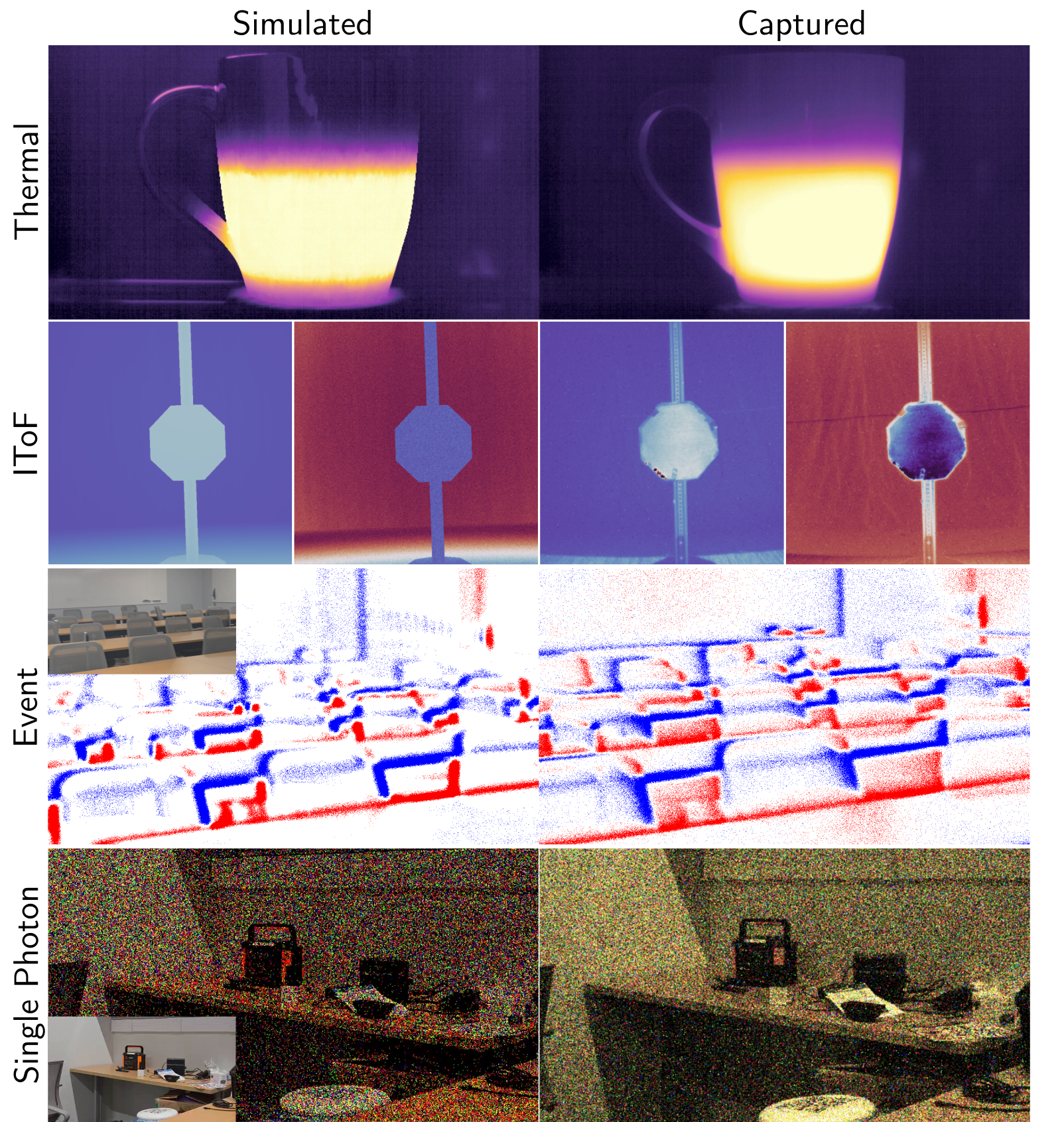}
    \vspace{-1em}
    \caption{\small{\textbf{Simulation Fidelity.} Comparison of simulated vs. captured scenes. The top two rows are simulated end-to-end using a digital twin of the real scene, while the bottom two emulate the corresponding sensors from a co-located conventional camera capture (inset).}}
    \label{fig:qual}
    \vspace{-1.5em}
\end{wrapfigure}

In the previous section, we calibrated a dToF sensor twin using a handful of real measurements. Here, in Fig.~\ref{fig:qual}, we further compare real and simulated sensor measurements for four additional modalities. The first two rows show sensor measurements simulated end-to-end using a digital twin of a real scene, and compared against captures of the same physical scene. The latter two emulate the simulated measurements using real conventional video, also validating our emulation protocol.

For a mug filling with hot tea, the simulated heating profile tracks a real-world capture from a FLIR Boson to within a mean absolute error of $7.6\%$ of the real mug's heating dynamic range over the 4 seconds following the fill. Against a real capture of a STOP sign taken with a Chronoptics iToF camera, our emulated iToF sensor achieves a mean phase error of $0.13$ radians across a $[0, 2\pi]$ range. At a 40 MHz modulation frequency, this translates to a mean depth error of about $5.3$ cm. Event counts emulated on RGB video of an auditorium differ from a real recording taken with a Prophesee EVK4 sensor by $1.5\%$ for positive events and $5.8\%$ for negative events. The photon detection statistics of our emulated Pi-Imaging SPAD sensor, likewise driven by RGB video, reach $0.221$ against a measured $0.229$. 

Small discrepancies remain in each case. The thermal error follows the emitters' surface emissivity and the way our heat-transport solver couples the mug to its surroundings: since the mug exchanges heat with a room whose ambient temperature we do not measure, a small error in either term propagates into the simulated surface temperature. For iToF, some of the discrepancy comes from the sensor's departure from the ideal. We assume a sinusoidal coding scheme, so the correlation model misses the multipath and higher-harmonic distortion that the physical Chronoptics device introduces. For the two modalities emulated from RGB video, discrepancies can arise before the sensor model runs. Our event-camera model matches the expected spread of contrast thresholds, but its front-end bandwidth and refractory period use the generic DVS model rather than the EVK4, potentially affecting event counts. The Pi-Imaging emulation faces a tougher challenge: mapping linearly rectified intensities to photon detection statistics requires knowing the source camera's unknown white balance and uncalibrated photometric response. As a result, the single-photon outcome is limited by the source camera, not our sensor model—an error that does not affect the digital-twin rows of Fig.~\ref{fig:qual}, which are simulated end-to-end.

\section{The \lib{} Multi-Modal Sensor Dataset}
\label{sec:dataset}
\vspace{-1.0em}
A key capability of \lib{} is \emph{multi-modal dataset generation}: a user selects scenes, camera trajectories, resolution and frame rate, the sensors to emulate, and their physical parameters, and the platform produces synchronized multi-sensor measurements with dense ground truth for a given set of tasks. To demonstrate this capability, we generate and release one such dataset, described below.
 
\noindent\textbf{An indoor instantiation.}
We curate $100$ unique indoor scenes spanning offices, kitchens, bathrooms, etc., each with a predefined $24$-second camera trajectory. Each scene is rendered at $100$fps in HDR at $1000\times1000$ as a stereo pair, with ground-truth poses, intrinsics, depth, surface normals, optical and scene flow, segmentation, temperature, and direct/indirect illumination components, yielding $480{,}000$ synchronized ground-truth views. From these, we emulate the full sensor suite: a single-photon camera at $16$\,kHz (via $160\times$ interpolation), a $4$\,kHz event camera ($40\times$), a $25$\,fps conventional camera, dToF and iToF LiDAR, a thermal sensor, an IMU at $250$\,Hz, and a stereo rig, using representative noise parameters drawn from the literature~\cite{SPADnoiseAnalysis,photocurrent-in-dvs,contrast-vs,Hu2021-v2e,GrossbergNayar_PAMI04,cmos-noise,vio-nonlin,spring,Gupta_2019b,GuptaToG_2018,MPD}. 

Simulating these streams guarantees perfect spatial co-registration and temporal synchronization—conditions that are exceedingly difficult to meet on a physical rig. Moreover, specialized hardware like high-resolution SPADs remains rare and expensive. As Table~\ref{tab:dataset_comparison} shows, our dataset provides broader modality and annotation coverage than prior datasets, with \emph{raw} sensor data (e.g., transient histograms) rather than just processed outputs.

\begin{table*}[t]
\centering
\scriptsize
\setlength{\tabcolsep}{3pt}
\renewcommand{\arraystretch}{1.24}
\begin{tabularx}{\textwidth}{
  |X  
  c   | c c c c  | c c c c c c c |}

\cline{3-13}
\multicolumn{1}{X}{} & & \multicolumn{4}{c|}{\textbf{Annotations}} & \multicolumn{7}{c|}{\textbf{Sensors}} \\[-1.5pt]

\hline
 
\textbf{Dataset} & {\textbf{Frames}} & \textbf{Depth} & \textbf{Flow} & \textbf{Sem.} & \textbf{Pt.\ Maps} & \textbf{Camera}  & \textbf{SPAD} & \textbf{Event} & \textbf{dToF} & \textbf{iToF} & \textbf{Therm.} & \textbf{IMU} \\
\hline
\rowcolor{adoptEstablished}
nuScenes ~\cite{nuscenes2019}~{\tiny\faTree} & 1.4M  & \faIcon{star-half-alt} & - & \faIcon{star}& \faIcon{star}&\faIcon{images} & - & - & \faCloud & - & -& \faCheck\\ 
\rowcolor{adoptEstablished}
SPICE-HL3 ~\cite{SPICE-HL3}{\tiny\faMeteor} & 1.2M  & - & - & - & - & \faIcon{images}&\faCheck& - & - & - & - & \faCheck\\
\rowcolor{adoptEstablished}
CroMo~\cite{cromo}~ {\tiny\faHome\faTree} & 80k & \faIcon{star-half-alt} & - & - & - & \faIcon{images} & - & - & - & \faCheck & - & \faCheck\\
\rowcolor{adoptEstablished}
MPD~\cite{MPD}~ {\tiny\faTree} & 103k  & - & - & \faIcon{star-half-alt} & - & \faIcon{camera} & - & - & - & - & \faCheck & -\\

\rowcolor{adoptRamping}
CARLA-Loc ~\cite{CARLA-LoC}~{\tiny\faTree} & 66k\rlap{*}  & \faIcon{star} & - & \faIcon{star} & \faIcon{star} & \faIcon{images} &- & \faIcon{images} & \faCloud & - & - & \faCheck\\
\rowcolor{adoptRamping}
TartanAir~\cite{Wang_2020}~{\tiny\faHome\faTree}   & 1M & \faIcon{star} & \faIcon[regular]{star} & \faIcon{star}        &\faIcon[regular]{star} & \faIcon{images} & - & - & \faCloud & - & - & \faCheck\\
\rowcolor{adoptRamping}
MSE~\cite{MSE}~ {\tiny\faHome}  & 100k & \faIcon{star} & - & \faIcon{star} & - & \faIcon{camera} & - & - & \faIcon{chart-bar} & - & - & -\\
\rowcolor{adoptRamping}
ThermVision-DB~\cite{thermvisionDB}~ {\tiny\faTree}  & 30k  & - & - & \faIcon[regular]{star} & - & - & - & - & - & - & \faCheck & -\\
\rowcolor{adoptRamping}
FLAT~\cite{FLAT}~ {\tiny\faHome} & 2k  & \faIcon{star} & - & - & - & - & - & - & - & \faCheck & - & -\\

\hline

\textbf{\lib{} (Ours)}~{\tiny\faHome}  & 480k   & \faIcon{star} & \faIcon{star} &\faIcon{star} & \faIcon{star} & \faIcon{images} & \faCheck & \faIcon{images} & \faIcon{chart-bar} &\faCheck & \faCheck & \faCheck \\
\hline
\end{tabularx}
\vspace{1pt}
{
\fontsize{6pt}{1pt}\selectfont
\colorbox{adoptEstablished}{\strut Real} \,
\colorbox{adoptRamping}{\strut Synthetic} \,
\textbf{Scene: }
\faIcon{home} Indoor\,
\faIcon{tree} Outdoor\,
\faIcon{meteor} Space \quad
\textbf{Streams: }
*Unsynchronized \quad
\textbf{Fidelity: }
\faIcon{star} High \,
\faIcon{star-half-alt} Med \,
\faIcon[regular]{star} Low \quad
\\[-1.5pt]\textbf{Views: } \faIcon{images} Multi-view \, \faIcon{camera} Single View \quad
\textbf{Format: } \faIcon{cloud}\, Point Cloud \, \faIcon{chart-bar} Transient Histograms}
\vspace{-4pt}
\caption{\small{\textbf{Landscape of computer vision datasets.} Existing datasets focus primarily on specific modalities (e.g., event cameras or LiDAR) or lack advanced sensor emulation (e.g., SPADs or thermal), \lib{} provides a comprehensive suite of multi-modal data including HDR stereo views, optical flow, point maps, and a wide variety of physically-based sensor emulations in a single unified dataset.}}
\label{tab:dataset_comparison}
\vspace{-1.75em}
\end{table*}

\noindent\textbf{Release.} We release the ground-truth components together with the scripts, assets, and checksums needed to regenerate the full sensor data (due to the large data size). The same scripts that regenerate our dataset let a user reconfigure it --- changing resolution, frame rate, the set of emulated sensors, or their parameters --- to produce a dataset tailored to their application. Implementation details of the generation interface are given in Appendix~\ref{app:dataset}.

\vspace{-0.6em}
\section{Limitations and Outlook}
\label{sec:outlook}

\vspace{-0.5em}
\lib{} aims to model the raw measurement formation of a broad range of conventional and emerging modalities, together with their characteristic noise and failure modes. The goal is to narrow the gap between idealized synthetic data and real sensor measurements, and support sensor-aware design and analysis that idealized simulators cannot. 
 
\noindent\textbf{Scope of the current results.} The digital twin case-study is demonstrated on a single commercial device and a small set of scenes; broader validation across devices, scenes, and operating conditions remains future work. The sensor-suite study is a preliminary, single-task case study (object detection under a CMOS-plus-event versus SPAD-plus-IMU comparison) rather than a comprehensive design analysis. All current evaluations are open-loop: we do not yet close the perception–control loop on a physical or simulated robot. Finally, the released dataset is indoor-only; adding outdoor scenes across various operating regimes is an important next step. 
 
\noindent\textbf{Architectural limitations.} Decoupling world simulation from sensor emulation requires an intermediate abstraction (e.g., depth and albedo for dToF). This enables efficient emulation of sensor-level effects such as pileup, dead time, and jitter, and provides a favorable compromise between speed, modularity, and physical accuracy. However, it precludes direct emulation of volumetric effects and higher-order light transport, as required for modeling complex weather phenomena. Because world simulation is ray-tracing-based, wave-propagation effects such as diffraction are also difficult to capture~\cite{Keller:62}, complicating extension to modalities such as mmWave radar. %
 
\noindent\textbf{Outlook.} A future direction is to integrate \lib{} with real-time engines such as Unreal Engine~\cite{unrealengine}, which could reduce rendering cost and broaden material and asset variety, while moving toward closed-loop and eventually real-time simulation. Incorporating wave-based propagation would open the door to radar and other diffraction-limited sensors, and a library of calibrated sensor presets would let widely-deployed devices be emulated out of the box. We see \lib{} as a step toward making sensor selection and perception-system design an evidence-based discipline, and we release the platform, assets, and generation scripts toward that goal.

\clearpage
\acknowledgments{This research was supported in parts by NSF CAREER Award 1943149, NSF grants CNS-2107060, 2107236 and 2513219, the NSF-NIFA AI Institute for Resilient Agriculture, NRT Award Number 2152163, ONR grant N000142412155, and Wisconsin Alumni Research Foundation via a Research Forward Initiative. We would additionally like to thank Varun Sundar and Sungjin Cheong for their help with data collection, and Mani Ramanagopal for insightful discussions on thermal simulation.}

\bibliography{references}  %
\newpage

\renewcommand{\figurename}{Supplementary Figure}
\renewcommand{\figurename}{Suppl. Fig.}
 \renewcommand{\thesection}{S.\arabic{section}}
\renewcommand{\theequation}{S\arabic{equation}}
\renewcommand{\theHsection}{S.\arabic{section}}
\renewcommand{\theHfigure}{S\arabic{figure}}
\renewcommand{\theHequation}{S\arabic{equation}}
\renewcommand{\theHtable}{S\arabic{table}}
\setcounter{figure}{0}
\setcounter{section}{0}
\setcounter{equation}{0}
\setcounter{page}{1}

\begin{center}
	\Large \textbf{\lib{}: Beyond Idealized Sensors — A Multi-Modal Simulator for Robot Perception}\\
	\Large \textit{Supplementary Material}\\
    \vspace{1em}
\end{center}

\renewcommand{\theHsection}{appendix.\thesection}

\section{Architecture}
\label{sec:methods}

\lib{} is designed from the ground up to be maximally flexible. A key design principle is its multi-layered architecture that enables it to simulate a variety of seemingly unrelated sensor modalities (including unconventional and emerging sensors such as single-photon cameras which excel in high-speed and high dynamic range environments~\cite{qbp,Liu_2022_WACV_extremehdr}) with minimal effort by reusing its different components and extending them with plug-ins if needed.

The simulation engine is made up of layers that are accessible both (a) as a command-line interface tool, which is useful to complete common operations such as rendering different types of ground truth data or emulating conventional cameras, and (b) as a full-fledged library, which enables more complicated workflows such as simulating multi-camera rigs. These layers are as follows:

\noindent\textbf{World Simulation Layer:} Generate rich ground truth data annotations using high-quality rendering engines, optionally interpolate to higher frame rates to emulate realistic motion blur and high-speed sensors.

\noindent\textbf{Sensor Emulation Layer:} Emulate different sensor modalities using realistic image-formation models, enabling physically accurate sensor noise. 

Once simulated, new datasets are stored with all relevant metadata in a way that enables easy iteration and random access, enabling easy interoperability with existing frameworks. 

While the default workflow is to use these layers in sequence to manipulate the data until it is in the desired format, these can be used individually and even out of order to support different applications. For instance, if we skip the simulation layer and instead provide real high-speed video captures (\emph{e.g.} from XVFI~\cite{sim2021xvfi}), we can emulate high-speed sensing modalities (\emph{e.g.} event cameras and single-photon cameras), or conventional cameras from this dataset using the sensor emulation layer. 

\begin{figure*}[t]
  \begin{center}
    \includegraphics[width=1.0\textwidth]{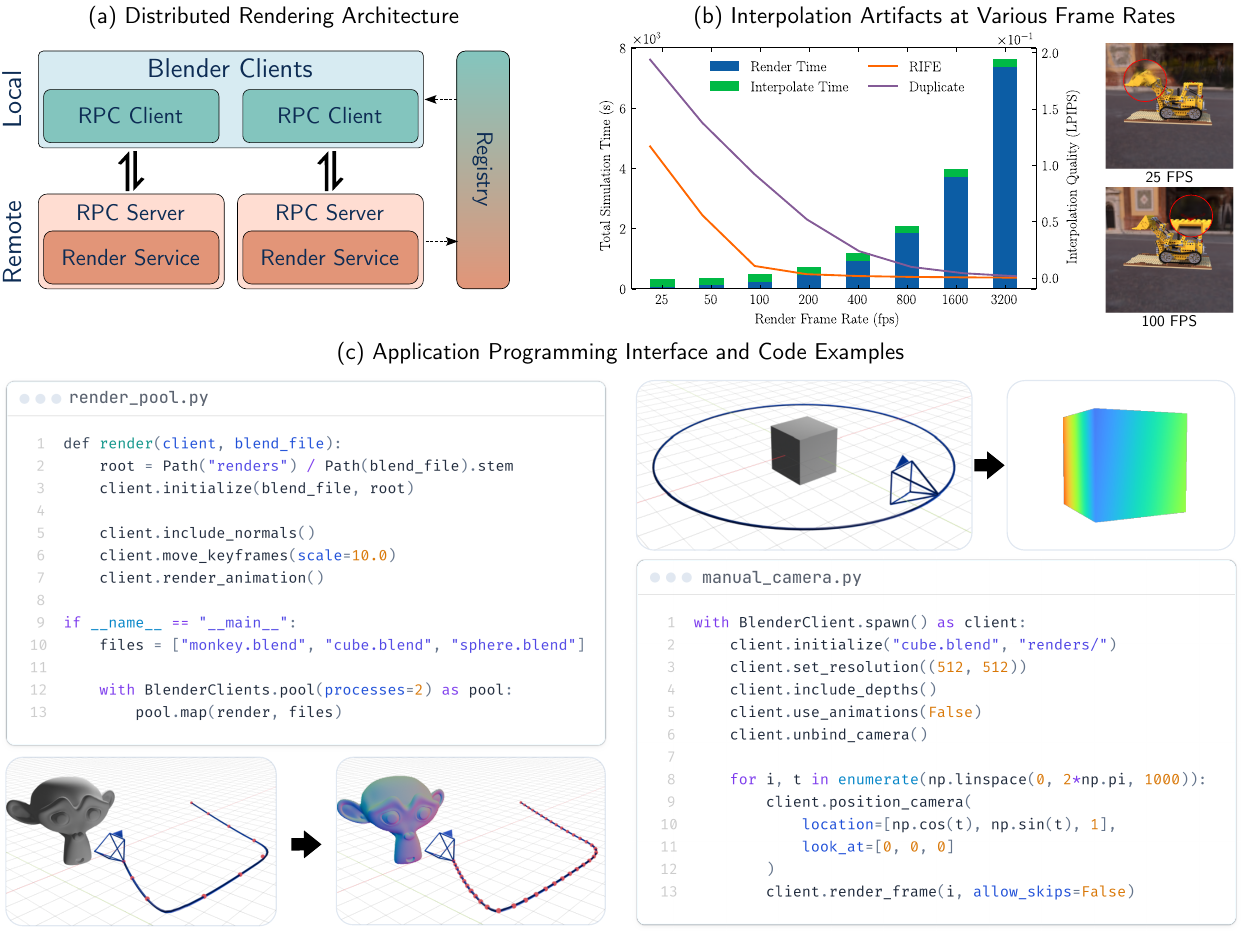}
  \end{center}
  \vspace{-1em}
  \caption{\textbf{Scalable Rendering and Interpolation:} (\textbf{A}) Client/server rendering architecture: multiple rendering services can be spun up on local or remote machines, each exposing a high-level rendering API via remote procedure calls (RPCs) and broadcasting their addresses over UDP for automated discovery via a registry server. This enables multiple clients to be orchestrated in tandem to distribute rendering jobs and speed up processing. (\textbf{B}) Interpolation quality/speed trade-off: we render a scene at different frame rates and interpolate using RIFE~\protect\cite{huang2022rife} or simple frame duplication until we reach $6400$ fps. A low rendering rate can introduce severe artifacts, shown here quantitatively (left) as a high perceptual similarity score (lower is better) and qualitatively (right), while a fast rate wastes compute for marginal quality gains. (\textbf{C}) Create a pool of two render workers and distribute work amongst them. For each of the three blender files, we enable the rendering of normals (L5), slow down the animations by a factor of ten (L6), and then render everything. By instead freeing the camera from any constraints we can move it manually and generate any arbitrary view. Here we spawn and connect to a single render client and fly the camera in a circle around the origin while always looking towards the origin. We also change the resolution (L3), render out depth buffers (L4), and disable any animations (L5).}
  \label{fig:rpc-interp}
\end{figure*}

\section{World simulation}
\label{app:world-sim}

The first step is to render out clean, blur and noise-free RGB images, depth maps, as well as normal maps, and other data modalities where applicable, which will both enable the emulation of different sensor types as well as act as ground truth. 

While there are many high-quality rendering engines such as Unreal Engine~\cite{unrealengine} or Mitsuba3~\cite{Mitsuba3}, we opt to use Blender~\cite{blender} as our primary rendering engine as it is easy to integrate with, widely accessible, well supported and produces photorealistic images. 

However, Blender's built-in API is very complex and not tailored towards high frame-rate rendering. It has a built-in maximum frame rate -- which is not sufficient to model event cameras or single-photon cameras natively, the latter of which can operate at hundreds of thousands of frames per second -- and has no easy way to output some ground truth annotations such as camera intrinsics. 

Further, rendering can be very costly and this cost scales with frame rate. By default, when a frame is rendered system utilization can be near a hundred percent, yet in between frames, or when setting up the scene for the next render, utilization is nearly zero, leading to spiky system load and underutilized resources further compounding slow render times. Finally, authoring new assets that target high fps is difficult as graphic designers are not used to doing so, and most existing assets would either playback at unrealistic speeds or be completely broken if they relied on more complicated effects such as geometry nodes or custom shaders. 

To address these challenges, we introduce a streamlined rendering API that simplifies the dataset generation workflow, makes it easy to reuse or create assets that have been authored for slower conventional frame rates, and allows for fast, distributed rendering on heterogeneous hardware.

\smallskip
\subsection{Streamlined API}
Blender provides an extensive python API, namely \texttt{bpy}, that exposes virtually all of Blender's functionality. However, it is only accessible through Blender's internal Python distribution or as a Python module if the user compiles Blender from its source. We instead provide a simplified API which provides a higher-level interface that streamlines the dataset generation process and automatically handles common pitfalls users might encounter.

This simplified API makes no assumptions about the scene that is loaded and will by default attempt to use all render settings the user set within Blender. While a new rendering pass, along with all necessary compositing nodes, might be automatically added to support different ground truth annotations, it is the user's responsibility to provide functional blend files. In Fig.~\ref{fig:rpc-interp} (C, left), multiple animations are rendered, and a rendering pass to get surface normals is automatically added if needed. However, an animation will not be rendered if its corresponding blender file is malformed in such a way that manually rendering a frame from within Blender does not work. Further, this simplified API makes it easy for users to generate different types of ground truth annotations and change common rendering settings such as targeting a different render device or changing resolution, as well as more advanced operations like manual camera updates within the render loop. 

\smallskip
\subsection{Distributed rendering}
To enable scalable simulations, we employ a client/server architecture, shown in Fig.~\ref{fig:rpc-interp} (A), that uses remote procedure calls (RPCs) as the communication medium. A few minimal dependencies are injected within Blender's runtime, which enables it to run as a server, listening for requests to set up scenes, move around cameras, render data, etc. Specifically, doing so from within Blender's runtime enables us to access all features through the \texttt{bpy} package and expose them via RPCs. 

Correctly orchestrating many rendering servers is technically challenging. First, to enable easy discovery of servers, a registry server is spawned and listens for UDP-based broadcast messages from rendered servers that are available. This rendezvous server communicates with the main client to inform it about available workers, enabling the user to connect to the server instances and distribute work amongst the worker pool, or even use multiple render servers to render out a single animation. Servers manage their own logging and exception handling, and, in the event of some unexpected crash, we ensure that the dataset is not corrupted and the process resources are freed.

This client/server architecture can drastically cut down on processing time and enables the use of a distributed set of machines with heterogeneous hardware. Even when worker processes are on the same machine, as is the case when using \texttt{BlenderClients}' \texttt{spawn} or \texttt{pool} directives (Fig.~\ref{fig:rpc-interp} (C)), render times can be shorter as the system's resources will be better utilized. Such distributed capability is a novel contribution to the Blender ecosystem as it provides the tooling for scalable and distributed renders -- which many companies, render farms, and add-ons provide as a paid service -- as a free and open-source library, which will democratize scaling of simulation-based bootstrapping of ML models.

For even larger-scale dataset generation tasks such as the dataset in Section~\ref{sec:dataset}, we employ a second layer of distributed computing through the use of HTCondor~\cite{chtc}, a high-throughput computing framework widely deployed in academic and institutional clusters. This involves containerizing our entire simulation stack, including all scripts, dependencies, and assets, using Docker~\cite{merkel2014docker}, which is automated in CI/CD, and submitting jobs to an HTCondor access point. This setup allowed us to leverage thousands of CPU cores and hundreds of GPUs across thousands of machines to render out the \lib{} dataset described in Sec.~\ref{sec:dataset}. 

\smallskip
\subsection{Keyframe remapping}
To address the challenges related to authoring high-speed assets and overcome Blender's maximum frame rate limitations, we propose to dynamically remap keyframes. In effect, keyframes are the control points that define the splines that govern all animations and dynamics, and, by remapping them, we can dilate time as needed. This enables easy re-use of existing virtual scenes and ensures that new frames do not break any pre-existing animations, as they are simply interpolated between keyframes. An example of this can be seen in Fig.~\ref{fig:rpc-interp} (C, left).  

\smallskip
\subsection{Physics engine}
Importantly, the keyframe-remapping technique enables us to take full advantage of Blender's powerful physics simulation engine. While it is slower than others~\cite{issacgym,Genesis}, the Bullet~\cite{bullet} physics engine is extremely powerful, especially since it can be customized using Blender's flexible geometry nodes, enabling users to build complex physical systems procedurally. Effects like fluid and cloth simulations, soft or rigid body physics, and more can simply be baked as keyframes and then remapped to any arbitrary frame rate. This enables us to easily create Flying Chairs~\cite{flyingchairs}-like datasets at much higher frame rates without needing to explicitly tweak the physics sub-time-stepping or collision engine. This is also the same physics engine used by Kubric~\cite{kubric}, although they manually integrate it into their pipeline using Bullet's Python API, restricting simulations to only rigid body physics.

\providecommand{\Tamb}{T_\mathrm{amb}}
\providecommand{\bn}{\hat{\mathbf{n}}}
\providecommand{\stefan}{\sigma}
\providecommand{\Eirr}{E}
\providecommand{\absorp}{\beta}
\providecommand{\Tstep}[1]{T^{#1}}

\subsection{Heat Transport Simulation}
\label{sec:world-heat-transport}
Several sensor modalities of interest---most notably long-wave infrared (LWIR, $8$--$14\,\mu\mathrm{m}$) thermal cameras---respond not to the instantaneous radiance leaving a surface, but to the long-wave radiation it \emph{emits} as a consequence of its temperature, and that temperature is the accumulated result of every watt of power the surface has absorbed over the preceding seconds to hours. Synthesizing a thermal image therefore requires synthesizing a plausible \emph{temperature field} first, which the sensor emulation layer (Sec.~\ref{sec:thermal-rendering}) then renders into a measurement. To support these modalities, VisionSIM integrates a fast, GPU-accelerated heat-transport solver directly into the world simulation layer, reusing the same Blender asset-authoring and rendering pipeline described above so that the temperature field is always consistent with the exact lighting the scene was authored with. This section develops the forward model in two stages: how absorbed visible light becomes a heat source (Sec.~\ref{sec:thermal-light-to-heat}), and how the resulting heat-transport PDE is discretized and solved (Sec.~\ref{sec:thermal-heat-transport}).

\subsubsection{Visible Light to Thermal Heat Source}
\label{sec:thermal-light-to-heat}
The bridge between the visible-band light transport that VisionSIM already simulates and the thermal domain is energy conservation at an opaque surface. The pixel value an ordinary camera records is the fraction of incident light \emph{reflected} toward it; since an opaque surface transmits nothing, the remaining energy is \emph{absorbed}, converted to heat, and subsequently redistributed by conduction, convection, and re-radiation, governed by the heat-transport equation~\cite{narayanan2025shape}. Reflection and absorption are thus complementary~\cite{maniCVPR24}, and the same path-traced irradiance the renderer evaluates to form the visible image also determines the heat deposited at each surface. Making this precise for a Lambertian scene point $\mathbf{x}$, let $\Eirr(\mathbf{x},t)$ denote the incident visible-band radiant flux density ($\mathrm{W}/\mathrm{m}^2$) arriving at $\mathbf{x}$ under the scene's lighting. A camera focused at $\mathbf{x}$ measures only the reflected portion, giving the visible image intensity
\begin{equation}
    I_v(\mathbf{p}) \;=\; \frac{a(\mathbf{x})}{\pi}\,\Eirr(\mathbf{x},t),
    \label{eq:visible-image}
\end{equation}
where $\mathbf{p}$ is the pixel imaging $\mathbf{x}$ and $a(\mathbf{x})\in[0,1]$ is the surface albedo. The complementary absorbed power per unit area---the quantity that drives the heat equation---is the remaining fraction,
\begin{equation}
    q_\mathrm{abs}(\mathbf{x},t)
    \;=\; \absorp(\mathbf{x})\,\Eirr(\mathbf{x},t),
    \qquad \absorp(\mathbf{x}) = 1-a(\mathbf{x}),
    \label{eq:absorbed-flux}
\end{equation}
which enters the thermal problem as a Neumann boundary source (Sec.~\ref{sec:thermal-heat-transport}). The broadband absorptivity $\absorp(\mathbf{x})\in[0,1]$ generalizes the ideal Lambertian factor $1-a$ to account for surface finish and angular reflectance, and reduces to it when the user ties absorption directly to the visible albedo.

The decisive modeling choice is that we obtain $\Eirr$ directly from the renderer rather than asking the user to specify a heat flux by hand, as classical FEM packages (ANSYS~\cite{ansys}, COMSOL~\cite{comsol}, OpenFOAM~\cite{openfoam}) require. Concretely, we read $\Eirr$ from the renderer's \emph{diffuse} irradiance passes---the direct and indirect diffuse light transport that already produces the matte component of the RGB image, including indirect bounces, environment lighting, sky and sun, and animated emitters---baked into a per-texel irradiance map at each timestep and resampled to mesh vertices by barycentric interpolation. The practical consequence is that the temperature field is always consistent with the exact lighting the scene was authored with: moving the sun to a different time of day moves the heat input, adding an emitter adds a heat source, and turning off a lamp removes its thermal contribution---all without the user authoring any thermal boundary condition.

\subsubsection{Finite Element Heat Transport Simulation}
\label{sec:thermal-heat-transport}

\paragraph{Governing PDE.}
We model heat transport in a solid domain $\Omega\subset\mathbb{R}^3$ (following \cite{narayanan2025shape}) with boundary $\partial\Omega$, exchanging energy with an ambient at temperature $\Tamb$ through convection and radiation while absorbing externally driven irradiance on $\partial\Omega$. The temperature field $T(\mathbf{x},t)$ satisfies the inhomogeneous heat equation
\begin{equation}
    \rho\,c_p\,\frac{\partial T}{\partial t} \;=\; \nabla\cdot(k\,\nabla T) + \dot{q}, \qquad \mathbf{x}\in\Omega,
    \label{eq:heat-pde}
\end{equation}
where $\rho(\mathbf{x})$, $c_p(\mathbf{x})$, and $k(\mathbf{x})$ are the spatially varying density, specific heat capacity, and thermal conductivity, and $\dot q$ is a volumetric source. We take $\dot q\equiv 0$ and inject all energy through the boundary. The transient response is governed by the thermal diffusivity $\alpha=k/(\rho c_p)$, which is the natural parameter to reason about: aluminum ($\alpha\approx 97\,\mathrm{mm}^2/\mathrm{s}$) diffuses heat roughly $500\times$ faster than glass ($\alpha\approx 0.5\,\mathrm{mm}^2/\mathrm{s}$), and it is this ratio rather than the conductivity alone that distinguishes their thermal appearance over the timescale of a video.

\paragraph{Boundary conditions.}
A surface has no thickness and hence zero mass, so it cannot store energy and any heat entering it must exit on the other side. On $\partial\Omega$ this balance is the Robin-type condition
\begin{equation}
    -k\,\nabla T\cdot\bn \;=\; \underbrace{\stefan\,\epsilon\,(T^4-\Tamb^4)}_{\text{radiation}} \;+\; \underbrace{h_c\,(T-\Tamb)}_{\text{convection}}, \qquad \mathbf{x}\in\partial\Omega,
    \label{eq:bc-robin}
\end{equation}
where $\bn$ is the outward unit normal, $\stefan$ is the Stefan--Boltzmann constant, $\epsilon\in[0,1]$ is the LWIR emissivity, and $h_c$ is the convective heat-transfer coefficient. The left-hand side is the conductive flux through the surface given by Fourier's law; the right-hand side is the flux lost to radiation and convection. Surfaces with externally imposed temperatures (a poured hot liquid, a thermostatic heater, a contact patch) are instead handled with Dirichlet conditions $T(\mathbf{x},t)=T_D(\mathbf{x},t)$ on user-tagged sub-regions $\Gamma_D\subset\partial\Omega$, and the field is initialized from a per-vertex attribute $T(\mathbf{x},0)=T_0(\mathbf{x})$ that defaults to $\Tamb$.

\paragraph{Combined heat equation.}
Following~\cite{narayanan2025shape}, we fold the boundary condition directly into the energy balance of a finite element exposed to the surface with area $A$. Adding the absorbed irradiance as an external input gives a single combined equation,
\begin{equation}
    \rho\,c_p\,\frac{\mathrm{d}T}{\mathrm{d}t}\,\mathrm{dv} \;=\; k\,\Delta T\,\mathrm{dv} + \mathbbm{1}_{\partial\Omega} \Bigl(\sigma\epsilon A\,(\Tamb^4-T^4) + h_c A\,(\Tamb-T) + q_\mathrm{abs} \Bigr)
    \label{eq:combined}
\end{equation}
where $\Delta$ is the Laplace--Beltrami operator, $ q_\mathrm{abs}$ is the input power per unit area ($\mathrm{W}/\mathrm{m}^2$) absorbed by the object on the boundary in the visible spectrum and $\mathbbm{1}_{\partial\Omega}$ is an indicator function which is 1 if the infinitesimal volume is on the surface and 0 otherwise.

\paragraph{Linearized radiation.}
The quartic in Eq.~\ref{eq:combined} makes the boundary term nonlinear in $T$. For smaller temperature differences the radiation term can be linearized without sufficient loss of accuracy \cite{narayanan2025shape},
\begin{equation}
    \sigma\epsilon\,(\Tamb^4-T^4) \;\approx\; 4\,\sigma\epsilon\,\Tamb^3\,(\Tamb-T).
    \label{eq:rad-lin}
\end{equation}
Crucially, this folds radiation into a \emph{constant} boundary stiffness term that references only the fixed $\Tamb$ rather than the evolving solution, so the system matrix can be assembled once and reused across the animation, with only the lighting-dependent right-hand side reassembled per frame. For high-temperature targets (welding, furnace interiors etc.) the radiation term can instead be re-linearized about the current solution each step, at the cost of per-frame matrix reassembly.

\paragraph{Spatial discretization.}
We represent the temperature as a piecewise-linear field over the discrete domain, $T(\mathbf{x},t)\approx\sum_i\varphi_i(\mathbf{x})\,T_i(t)$, with $\varphi_i$ the nodal basis functions, and discretize $\Omega$ into finite elements with $V$ vertices. The discrete Laplace--Beltrami operator is $\Delta_\Omega=M_v^{-1}L$, where $L\succeq 0$ is the cotangent Laplacian and $M_v$ the lumped vertex mass matrix; a second, area-weighted mass matrix $M_a$, assembled by summing one third of each boundary triangle's area into its vertices, integrates the boundary terms. Generic asset meshes are typically open, non-manifold, and full of degenerate triangles, which break a naive cotangent Laplacian. We therefore assemble $L$ with an intrinsically mollified cotangent operator~\cite{Sharp:2020:LNT}, which regularizes needle triangles, slivers, and near-coincident vertices so the system stays well-conditioned on geometry that no conventional FEM package would accept without remeshing.

\newcommand{\dtd}{t}
\paragraph{Time integration.}
Substituting the linearized radiation Eq.~\ref{eq:rad-lin} into the discretized combined equation Eq.~\ref{eq:combined} and applying implicit backward Euler, $\dot{\mathbf{T}}\approx(\Tstep{n+1}-\Tstep{n})/\dtd $, each timestep is a single sparse linear solve. Gathering $\Tstep{n+1}$ on the left and multiplying through by $M_v$ gives
\begin{equation}
\begin{aligned}
    \Big(M_v + \alpha\,\dtd\,L + \mathbbm{1}_{\partial\Omega}\tfrac{\dtd}{\rho c_p}\big(4\stefan\epsilon\Tamb^3 + h_c\big)M_a\Big)\Tstep{n+1} \;=\; M_v\,\Tstep{n} \\[2pt]
    {} + \mathbbm{1}_{\partial\Omega}\tfrac{\dtd}{\rho c_p}\big(4\stefan\epsilon\Tamb^4 + h_c\Tamb + q_\mathrm{abs}^{\,n+1}\big)M_a\,\mathbf{1},
\end{aligned}
    \label{eq:be-step}
\end{equation}
where $\mathbbm{1}_{\partial\Omega}$ is the boundary-vertex indicator and the per-vertex scalars $\epsilon,\rho,c_p,\beta, q_\mathrm{abs}^{\,n+1}$ are read from the gathered material vectors. The left-hand matrix is sparse, symmetric positive definite, and---because radiation is linearized about the fixed $\Tamb$---constant across the animation; only the right-hand side, through $ q_\mathrm{abs}^{\,n+1}$, changes when the lighting is animated. It is therefore prefactored once and the per-step solve is dramatically cheaper, warm-started from the previous timestep with a matrix-free conjugate-gradient iteration that keeps the iteration count low for smoothly evolving lighting.

\paragraph{Volumetric heat simulation.}
A surface-only solve suffices for thin shells (sheet metal, fabric, leaves), but it misrepresents the diffusion pattern for thicker objects, and it fails outright for the hollow meshes common in asset stores: a mug or a skillet modeled as a thin shell conducts heat as if it were a hollow surface rather than a solid body, producing visibly wrong behavior. One way to fix is by tetrahedralizing the interior, but watertight tetrahedralization is unworkable on the open, self-intersecting meshes these assets typically are. We instead work with points rather than tetrahedra. For each closed (or near-closed) object we sample a dense, blue-noise interior point cloud over the bounding box by Bridson Poisson-disk sampling~\cite{bridson2007sampling}, keeping only points that pass a BVH ray-parity inside test: a point is interior iff a random ray crosses the surface an odd number of times, and we take a $256$-ray majority vote for robustness to degenerate geometry. A mollified point-cloud Laplacian~\cite{Sharp:2020:LNT}, built on a $k$-nearest-neighbour graph over the combined surface-plus-interior point set, then couples interior diffusion to the boundary degrees of freedom exactly as a volume FEM would, but with no meshing required. In Eq.~\ref{eq:be-step} the indicator $\mathbbm{1}_{\partial\Omega}$ simply selects the original surface vertices: interior points diffuse but receive no irradiance, radiation, or convection. The resulting sparse systems are assembled and solved on the GPU, keeping the solve fast enough to run in the loop of a dataset-generation pipeline; per-vertex temperatures are written back as a standard Blender attribute that the downstream sensor emulation layer (Sec.~\ref{sec:thermal-rendering}) consumes.

\paragraph{Scope of the heat-transport model.}
Two modeling choices bound the fidelity of the temperature field. First, we read the absorbed power from the renderer's \emph{diffuse} irradiance passes (Eq.~\ref{eq:absorbed-flux}), neglecting the specular component of the incident radiance; since specularly reflected energy is reflected rather than absorbed, attributing it to $q_\mathrm{abs}$ causes our heat source to systematically \emph{over-estimate} the power deposited at glossy surfaces. Second, the mollified Laplacian~\cite{Sharp:2020:LNT} built over a $k$-nearest-neighbour point set is robust on the open, non-manifold meshes typical of asset stores, but this robustness comes at a cost: no discrete Laplacian can satisfy symmetry, locality, linear precision, and positivity simultaneously~\cite{Wardetzky2007}, and the point-cloud solve forgoes the conforming guarantees of a tetrahedral discretization, leaving its accuracy sensitive to point density and to the statistical ray-parity inside/outside test. Richer alternatives such as steady-state walk-on-spheres Monte Carlo estimators are a natural extension and could be plugged in behind the same interface in the future.

\smallskip
\subsection{Deriving Additional Annotations}

\paragraph{Disparity maps.}
For stereo image pairs, a disparity map can be computed directly from the ground truth depth map if the camera parameters and the baseline are known. Assuming a rectified stereo setup with a baseline $B$ and focal length $f_x$, the disparity $d$ at a pixel with depth $Z$ is simply given by $d = \frac{f_x \cdot B}{Z}$.

\paragraph{Point maps.}
Point maps, which store the $3$D coordinates $(X, Y, Z)$ of the scene point corresponding to each pixel, can also be derived from depth maps. These can either be obtained directly from the simulator---which leads to large file sizes but better precision due to fewer floating-point errors---or computed post-hoc by backprojecting the depth map. Specifically, given a pixel $(u, v)$, depth $Z$, and camera intrinsics with focal lengths $f_x, f_y$ and principal point $(c_x, c_y)$, the point coordinates in the camera's frame are $X_c = \frac{(u - c_x) Z}{f_x}$, $Y_c = \frac{(v - c_y) Z}{f_y}$, and $Z_c = Z$. Point maps are stored in the world frame, so this backprojection is followed by the rigid transform $[X, Y, Z]^\top = R [X_c, Y_c, Z_c]^\top + \mathbf{t}$, where $R$ and $\mathbf{t}$ are the camera's rotation and translation for that frame. Deriving point maps post-hoc therefore requires the pose, the camera intrinsics, and the depth map; the simulator's own point maps already apply this transform.

\paragraph{Scene flow.}
Scene flow can be roughly thought of as the $3$D analog to optical flow. Instead of the two-dimensional motion field mapping pixel coordinates from one frame to the next, it captures the displacement of scene points in world coordinates. 

By encoding scene dynamics, this three-dimensional vector field is particularly useful for navigation or robotics tasks, however, historically scene flow datasets have been underrepresented. One of the largest scene flow datasets is the Spring dataset~\cite{spring}, which relies on a custom patch of Blender to provide scene flow directly as a render pass.

While this is without a doubt the fastest way to get scene flow, as a by-product of rendering, we cannot make use of it in \lib{} as it would require the framework to either maintain a custom branch of Blender or force users to compile the patched version themselves. 

Instead, we opt to compute scene flow using other ground truth annotations which can all be generated using \lib{} without relying on a custom Blender patch. This significantly reduces the complexity associated with distributing \lib{}, but has two main downsides. 

First, scene flow generation is now a post-hoc process, to compute forward flow, the ground truth depth of the next frame needs to be known, and vice-versa for backward flow. Since this process is not causal it has to lag the main rendering process. Currently, \lib{} opts to compute scene flow exclusively after the render step is complete, to ensure dataset integrity in the event of an irrecoverable crash.    

Second, to generate ground truth flow vectors, Blender needs to know ahead of time where the state of the world at the next (and previous) timestep. The implication is that both optical and scene flow cannot be generated when the user manually positions the camera within the render loop as shown in Fig.~\ref{fig:rpc-interp}(C). Instead, the camera motion needs to be known pre-render, and communicated with Blender via the use of keyframes. In practice, this is seldom a limitation as \lib{} provides utilities to set camera keyframes before invoking \texttt{render\_animation}.  

To compute the forward scene flow $S_f$ from time $t$ to $t+1$ we need the point maps $PM_t$, $PM_{t+1}$ and forward optical flow $F_f$ for those times. We first get the world coordinates $(x_t, y_t, z_t)$ for a pixel at time $t$ using $PM_t$. Then, use $F_f$ to warp the pixel coordinates from frame $t$ to their corresponding locations in frame $t+1$, which yields $(u_{t+1}, v_{t+1})$. We sample $PM_{t+1}$ at this location and obtain the corresponding world coordinates $(x_{t+1}, y_{t+1}, z_{t+1})$. The difference between these two points in world coordinates is the desired forward scene flow.

\smallskip 
\subsection{Interpolation}
\label{app:interpolation}

To accurately model high-speed sensors and artifacts such as motion blur, high frame rate data might be needed. While \lib{}'s distributed architecture, equipped with its powerful keyframe remapping features, allows users to render datasets at arbitrary frame rates, in many cases, this is not needed. Every scene has a \textit{native frame rate}, beyond which any motion or view-dependent effect becomes imperceptible, wasting additional rendering efforts. 

In the real world, sampling the scene much faster than this intrinsic rate is beneficial. It reduces the observed noise and boosts the measured signal. This is especially important in the case of single-photon cameras, which can run at a hundred thousand frames per second, yet each observation encodes very little information — a single bit's worth in fact~\cite{SPADnoiseAnalysis}.

However, in simulation, we have access to perfect, noise-, and blur-free frames, so there is no need to render redundant data, it suffices to render \textit{fast enough} and interpolate the rest of the way to reach our desired frame rate. This technique, to render ground truth at some slower rate and fill in the rest, can dramatically cut down on render times for scenes that have low native frame rates. The minimum frame rate at which to render ground truth frames depends entirely on the scene, its geometry, material properties, textures, camera parameters and resolution, and the speed of the scene's relative motion. In general, the native frame rate is hard to estimate, even in simulations where all parameters are known. 

Fortunately, precisely estimating this inflection point is usually not needed. Overshooting means that there will be extra overhead as we are required to render a few additional frames that are near duplicates. On the other hand, if we render slower than the native frame rate, adjacent frames will be more dissimilar. Thankfully, modern deep learning-based video interpolation methods can help fill this gap as they are specifically designed to interpolate between frames that encode large motion. In the limit, when using these methods on frames that have been rendered at or past the native frame rate, adjacent frames are identical and interpolation becomes trivial, effectively devolving into frame duplication. Conversely, when frames are too far apart interpolation will result in noticeable artifacts. Therefore, these interpolation methods give us some leeway in finding the best trade-off between rendering more ground truth frames, at the cost of additional rendering time, and interpolating between wider gaps in the data, which can lead to artifacts. 

In Fig.~\ref{fig:rpc-interp} (B) we rendered a test trajectory at many different frame rates and in each case interpolated the resulting data up to a common frame rate of $6400$ fps. We plot the average perceptual similarity between these interpolated frames and ground truth data (lower is better), as well as the total wall time to render and interpolate these different datasets as measured using a single rendering job, on an Nvidia RTX 3080. Notice that the time it takes to interpolate the frames is dwarfed by rendering time. The overall quality of the data improves with higher frame rates and lower interpolation factors, hitting an inflection point around $200$fps, after which there are diminishing returns as rendering time explodes. From this, we can see that \textit{for this specific scene} the native frame rate is around two to four hundred frames per second, and below this rate we incur significant interpolation artifacts, as shown. It takes just under $20$ minutes to render this scene at $400$fps and interpolate it $16 \times$ to $6.4$ kHz, as compared to over four hours if we rendered every frame. 

In practice, one would like to determine the optimal keyframe multiplier $K$ that achieves this native frame rate automatically. Prior work such as ESIM~\cite{esim} addressed this by dynamically triggering a new render whenever the per-pixel optical flow exceeds a fixed threshold. While theoretically appealing, this strategy is difficult to parallelize efficiently and produces non-uniform temporal sampling that is incompatible with many video interpolation methods and sensor modalities. We instead propose a simple yet highly effective closed-form iterative approach. Under the first-order approximation that the optical flow magnitude $\mathcal{V}$ scales as $K^{-1}$ --- i.e., speeding up the frame rate by $K$ reduces frame-to-frame motion proportionally --- we can estimate the corrected multiplier $K_{i+1}$ needed to scale the flows to a target value $\mathcal{V}_{target}$:
\begin{equation}
    K_{i+1} = K_i + \alpha^i K_i \cdot\left(\frac{\hat{\mathcal{V}}_q(K_i)}{\mathcal{V}_{target}}-1\right)_{[a, b]} \,.
    \label{eq:keyframe_newton}
\end{equation}
where $\hat{\mathcal{V}}_q(K_i)$ is the measured $q$-th percentile flow magnitude at multiplier $K_i$, the subscript denotes clamping to the range $[a,b]$ to prevent over or under shooting, and $\alpha^i$ acts as a dampening factor, with $\alpha$ typically set to $0.95$. When there is no dampening or clamping, this equation reduces to directly scaling the keyframe multiplier by the multiplicative factor, which would --- if flow is perfectly inversely proportional to $K$ --- result in the target flow. The added terms help with convergence when this assumption breaks down.

In this way the frame multiplier is set by the scene complexity and speed, through the optical flow proxy, as opposed to by the sensor we wish to emulate. We render at a rate that keeps the 95th-percentile optical flow at a few pixels and interpolate the remaining frames, so the interpolated sequence stays within the tolerance plotted in Fig.~\ref{fig:rpc-interp}(B) of the fully rendered ground truth. Because the rate comes from the per-scene search below rather than a fixed multiplier, it adapts to each trajectory, and a single render supports emulation at any sensor frame rate above it.

To estimate $\hat{\mathcal{V}}_q(K_0)$, we need to render the optical flow of the full trajectory; however, this is much cheaper than rendering the full image frames. Not only can we speed up rendering by using a very low number of samples-per-pixel and disabling auxiliary passes, but we can also render at a proportionally smaller resolution and then scale up the flow fields accordingly. In practice, this technique converges to the desired operating point in just a few iterations, making it inexpensive to execute as a pre-processing pass before any full-scale rendering begins. As an example, on the scene in Fig~\ref{fig:rpc-interp} (B), this technique finds, in just two iterations, that to achieve a $95$-percentile flow of less than two pixels we should use a keyframe multiplier of $4.16$, meaning a native frame rate of $99.8$fps, perfectly matching up with the inflection point on the plot. Rendering at this frame rate and interpolating up to the target rate of $6400$fps reduces the total simulation time from 4 hours and 3 minutes to just under 4 minutes, a $60\times$ speedup. 

\begin{figure*}[t]
  \begin{center}
    \includegraphics[width=1.0\textwidth]{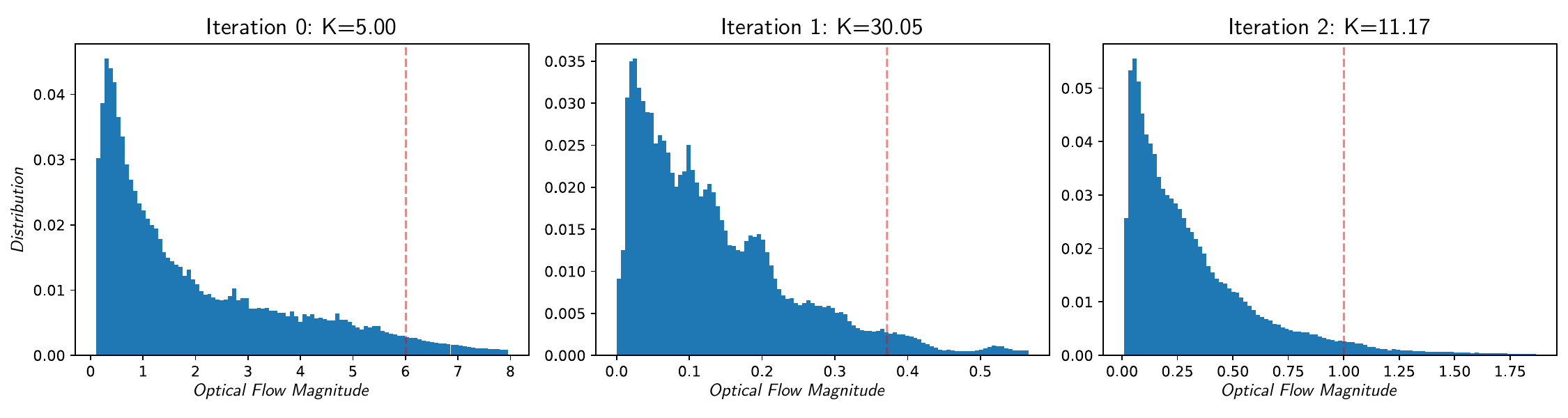}
  \end{center}
  \vspace{-1em}
  \caption{\textbf{Convergence of the keyframe-multiplier search on the \emph{classroom} scene.} Each histogram shows the distribution of per-pixel optical flow magnitudes measured at a different iteration, with the target flow of $1$\,px. The 95th-percentile estimate $\hat{\mathcal{V}}_{95}$ is shown as the dashed red line for each iteration. The first iteration overshoots, as many smaller flows are discovered to be around $\tildeNice0.2$px, while the second iteration finds the optimal keyframe multiplier. }
  \label{supfig:framerate-optim}
\end{figure*}

Fig.~\ref{supfig:framerate-optim} illustrates the convergence of this procedure on the \emph{classroom} scene~\cite{blenderexamples} with a target flow magnitude of $\mathcal{V}_{target}=1$\,px. Starting from an initial multiplier of $K_0 = 5$, the measured 95th-percentile flow is $\hat{\mathcal{V}}_{95} \approx 6$\,px --- well above target, indicating the frame rate is too low. Following Eq.~\ref{eq:keyframe_newton}, the multiplier is updated to $K_1 = 30$; however, the flow distribution shifts and concentrates near $0.2$\,px, yielding a 95th-percentile of only $0.37$\,px, meaning we have overshot the target. A final update produces $K_2 = 11.2$, for which the 95th-percentile flow lands precisely at the $1$\,px target.

Finally, while this method produces substantial speed-ups with little impact on the quality for natural images, it does not work for all modalities. The input images need to be in-distribution for video interpolation methods to work well. Specifically, for non-tone-mapped, high dynamic range images, common methods such as RIFE~\cite{huang2022rife} or FILM~\cite{reda2022film} do not work reliably. In these cases, we can either render every frame or, by also using ground truth depth maps, camera intrinsic, and pose, we can reproject pixels from one frame to a new one. This will cause a different type of artifact, namely bands of missing pixels due to parallax effects, which could be addressed using inpainting or colorization methods~\cite{telea,levin-colorization}.

\section{Physically accurate sensor emulation}
\label{app:sensors}

At this stage, we have access to high-speed interpolated data, which we will need to further process into the desired sensor modality. Specifically, now that world simulation is done, we perform sensor-realistic emulation for single-photon, high-speed, event, and conventional cameras, along with IMUs. Other camera types could be added by plugging in their image formation model on top of the simulated and interpolated data. 

\noindent\textbf{Conventional cameras:}
While rendered frames are good proxies for what a conventional camera might capture in ideal conditions, in many challenging real-world settings, these are significantly different, and accurate sensor modeling is required~\cite{sensortransfer,quantifyingsim2real}. 

The brightness of a conventional CMOS camera pixel can be modeled as a function of the scene radiance over the exposure time transformed by the camera's optical stack and response function $\Gamma$, which encapsulates the camera response curve, tone mapping, quantization, and proprietary image signal processing~\cite{GrossbergNayar_PAMI04}. Specifically, over an exposure time $\tau$, a CMOS pixel will accumulate charge, and potentially saturate once it reaches its full well capacity (FWC). This process is subject to many sources of noise, including read noise and photon shot noise, which we absorb into $\mathcal{N}$. The intensity $I$ of the pixel can be modeled as: 
\begin{equation}
    I = \Gamma \left(min\left(\int_{\tau} \phi dt,~\text{FWC}\right) + \mathcal{N} \right) \,.
    \label{eq:camera_model}
\end{equation}

To emulate this, we sum together many interpolated intensity frames, discretizing the above integral and creating motion blur. It is important that this discretization be performed at a high frame rate otherwise, the motion blur will not be smooth and will lead to noticeable ghosting artifacts. We then add Gaussian read noise, apply a gain factor representing the sensitivity (ISO), and quantize to a certain bit-depth: the sensor model is adopted from the work of Hasinoff et al. \cite{Hasinoff:2010:Noiseoptimal}. More complex effects, such as optical aberrations or fixed pattern noise, can be added here as well.

The camera's image signal processor (ISP) pipeline involves many computational steps after signal readout, such as de-mosaicing, de-noising, tone-mapping, and compression. The current version of the simulator provides baseline implementations for each of these steps (see methods). Future additions may include 3A algorithm implementations (auto-exposure, auto white-balance, and auto-focus), and recent learning-based imaging pipelines (``neural ISPs'') that are optimized end-to-end. 

\noindent\textbf{High-speed cameras:}
Usually, also based on CMOS technology, high-speed cameras can be emulated similarly to conventional cameras. However, these suffer from low signal, as the integration time is short, and readout noise that can be an order of magnitude larger than their conventional counterparts\footnote{For example Scimedia's MiCAM03-N256 has \href{https://www.scimedia.com/products/is/micam03-n256/}{readout noise of up to $130 e^-$}}. Thus, we can use the same model, and adapt $\tau$ and $\mathcal{N}$ accordingly.

\noindent\textbf{Stereo and multi-view cameras:}
Stereo cameras enable depth perception by capturing images from two viewpoints with a fixed baseline. We model stereo rigs by spawning two cameras with a fixed offset and intrinsically synchronizing their shutters. This allows for the generation of perfectly aligned stereo pairs and ground-truth disparity maps, which are essential for training and evaluating stereo matching algorithms~\cite{Middlebury2014,spring}. Multi-view cameras can be modeled similarly by spawning more cameras with different baselines and intrinsics.

\noindent\textbf{Passive single photon cameras:}
Despite being excellent general purpose imagers, capable of extreme high-speed, low-light, and high-dynamic range imaging~\cite{spc-demo}, single-photon cameras remain scarce and expensive. Emulating these cameras can help meet the need for cheap, high-quality, single-photon data and encourage future research.  

Specifically, we emulate a single-photon avalanche diode-based camera, as these are rapidly becoming mainstream thanks to their CMOS-compatible fabrication process~\cite{SwissSPAD2}. These are binary devices that do not suffer from read noise as they are entirely shot-noise-limited. However, SPC pixels reset after a photon detection event, during which no new photons can be detected.
This effectively quantized the pixel's output, leading it to follow a Bernoulli distribution parameterized by $1-e^{-\phi \tau}$~\cite{SPADnoiseAnalysis}, where $\phi$ is the scene's radiant flux. By using the high-speed interpolated frames as a proxy for scene flux, we can emulate an SPC by sampling this distribution. Doing so accurately involves rectifying the tone-mapped frames to their linear intensities and accounting for other non-idealities such as dark count rates or low fill factors. See the supplement for more.  

\noindent\textbf{Active single-photon cameras:}
Robust depth perception using direct time-of-flight (dToF) sensors requires a high-fidelity, physically accurate model of an active single-photon camera (ASPC), also commonly known as single-photon LiDAR. We simulate the complete acquisition pipeline, mapping scene irradiance and illumination conditions to noisy photon timestamps. 
For each pixel, scene points within the FoV are accumulated and mapped to time-of-flight delays to form the scene response function (SRF). Convolution with the instrument response function (IRF) then models the system's temporal blur, while ambient illumination is treated as a uniform background rate. Together, these define the per-pixel photon arrival rate, from which discrete photon arrivals are drawn as an inhomogeneous Poisson process. 

In practical single-photon LiDARs, not all photon arrivals are converted to photon detections due to sensor dead time.
We explicitly model the detector's dead time $\tau_{dead}$ using a non-paralyzable dead time model where photons arriving within the recovery period are discarded. Our implementation can easily be extended to support paralyzable dead time. These functionalities accurately reproduce pile-up distortion, which is essential for emulating physically accurate LiDAR measurements and validating robust estimation algorithms. In addition to simulating raw photon timestamps, the simulator supports both standard Equi-width histogramming (EWH) methods and more recent Equi-depth histogramming (EDH) methods \cite{count-free-spc, pedh}.

\noindent\textbf{Event cameras:}
By asynchronously emitting per-pixel brightness changes instead of full frames, event cameras are well-suited for many latency-critical applications such as object tracking~\cite{event-tracking,Messikommer23cvpr}. Our event model follows v2e~\cite{Hu2021-v2e}. Each pixel applies a probabilistic contrast threshold: a positive or negative event is emitted when the log-intensity change since the last event exceeds the threshold, with the threshold drawn stochastically per event to reproduce the threshold variation measured across a real sensor. The photodiode front end is modeled as a first-order low-pass filter, which sets the bandwidth limit that suppresses events under rapid illumination changes. A centre--surround spatial kernel applied to the pixel's input reproduces the DVS contrast response. After an event, the pixel is held for a refractory period during which further changes are not reported. These stages take the interpolated high-frame-rate intensity sequence as input, which \lib{} has already produced, so event emulation runs directly on both rendered and migrated video. 

\noindent\textbf{Inertial measurement units:}
Finally, we model IMUs as they are often co-located with many imaging sensors and are an integral part of many robotics and AR/XR applications where visual-inertial or SLAM-based algorithms may be used. These contain an accelerometer and gyroscope that are modeled by a discrete-time random walk~\cite{sigma-point-imu,vio-nonlin,forster2015imu}, which introduces both noise and a slow drift away from the camera's actual linear acceleration and angular velocity, respectively.
The reference values for both are estimated from the ground-truth camera pose sequence through standard numerical differentiation techniques.
The noise standard deviations are set by default to values obtained in previous works in visual-inertial odometry \cite{vio-nonlin}, but may also be set differently using the datasheet of the sensor to be emulated.

\noindent\textbf{Indirect time-of-flight LiDARs:}
Indirect time-of-flight (iToF) LiDARs estimate depth by measuring the phase shift or correlation between a modulated light source and the reflected signal~\cite{lange2001}. Unlike dToF sensors that measure individual photon timestamps, iToF sensors typically modulate the active illumination with a periodic coding function---such as a sinusoid, square wave, or more exotic functions~\cite{GuptaToG_2018}---and record the integrated intensity over several correlation windows. Our simulator supports many common coding functions, enabling the study of different modulation schemes in challenging environments.

\noindent\textbf{Thermal cameras:}
Unlike other modalities, long-wave infrared (LWIR) thermal cameras measure the radiation emitted by a surface due to its temperature. We emulate this by first rendering the temperature field produced by the world simulation's heat-transport solver into in-band thermal radiance using a physically motivated graybody emission model (Sec.~\ref{sec:thermal-graybody}). We then model the temporal dynamics of the dominant commercial LWIR detector, the microbolometer. By filtering the rendered radiance to account for the detector's thermal inertia, we accurately reproduce the characteristic lag and motion smear of real thermal video (Sec.~\ref{sec:thermal-microbolometer}).

\section{Data storage and ML framework integration}
\label{app:datastorage}
Once the desired data has been generated, it needs to be stored in a convenient format that enables fast random access and iteration and easily integrates into downstream frameworks. While many such formats exist, we extend the data format informally introduced by NeRFs~\cite{mildenhall2020nerf} to maintain compatibility with existing view synthesis and multi-view reconstruction frameworks such as NeRF-Studio~\cite{nerfstudio} or COLMAP~\cite{colmap}. Specifically, we extend this format along two fronts: a sensor-agnostic, and sensor-aware extension.   

Both are based around maintaining a \texttt{transforms.json} file which encodes details about the camera's intrinsic parameters, such as its resolution, focal length and principal point as well as a list of camera poses and asset paths for each instant of the camera trajectory. In its simplest form, these asset paths are just relative paths to the rendered image~\cite{mildenhall2020nerf}. This format can be extended in a sensor-agnostic way, by allowing these asset paths to be more complex to account for other sensor modalities and ground truth annotations. For instance, events are encoded as a stream of $(t, x, y, p)$ tuples encoding the time, x/y-coordinates, and polarity of the event, which are saved in a separate file that can be linked to. 

For single-photon camera data, a sensor-optimized scheme is more appropriate since saving individual single-photon frames can use enormous amounts of storage as image codecs are poorly suited at encoding such highly stochastic data~\cite{jungerman2024radiancefieldsphotons}. Further, many applications, such as Gaussian Splatting~\cite{kerbl3Dgaussians}, ingest data at the granularity of pixels instead of full frames, which would require decoding a whole binary frame to sample just a few pixels. This is not an issue when there are only a few hundred images as pixel values can be pre-loaded and cached on the GPU for fast training~\cite{mueller2022instant}, but we quickly run into memory limitations if we attempt to pre-load potentially millions of single photon frames. 

To overcome these issues, we store all SPC data as a single large bitpacked numpy array. In this manner, the data is losslessly compressed by $8 \times$ and can be memory mapped which enables fast random access to the whole dataset. Since the minimum addressable unit of memory is a byte, we are forced to read in a whole byte, whether it represents eight pixels or only one. Thus this compression scheme has virtually no runtime costs, and decompression can be performed on the fly when accessing pixels. \lib{} abstracts away the tricky bitmasking operations required to perform this compression, instead exposing a high-level API to read and write SPC data directly.

For both these formats, \lib{} provides data-loading utilities that can directly be used with modern deep learning frameworks such as PyTorch or TensorFlow to train machine learning models.

\section{Forward models and implementation details for selected sensors}

\subsection{Emulating single-photon cameras}
Assuming constant flux during the exposure time, the number of photons incident on a pixel during an exposure time $\tau$ follows a Poisson distribution parameterized by the total incident light during that exposure, namely $\phi\tau$, where $\phi$ is the scene's radiant flux. A SPAD-based SPC does not suffer from read noise; however, its pixels reset after a photon detection event, during which no new photons can be detected. This quantizes the pixel's output, leading it to follow a Bernoulli distribution with parameter $1-e^{-\phi \tau}$~\cite{SPADnoiseAnalysis}. 

Therefore, SPC can be emulated by Bernoulli sampling the interpolated intensity frames. However, the ground truth frames do not encode scene flux; they encode the scene's intensity, often non linearly. We must first ensure the frames are in a linear colorspace, either by generating them as such or undoing any tone mapping that may have been applied. This intensity then needs to be mapped to scene flux; however, up until now, there has been no concept of scene brightness in absolute terms. While a user can set the power of every light source in Blender, this does not translate directly to physical units. Instead, we introduce a correction factor $f$ that maps a pixel's linear intensity to a flux value, absorbing into it any non-idealities such as dark count rates or low fill factors. 

\subsection{Active single-photon camera modeling}
\label{subsec:aspc_modeling}

Robust depth perception using direct time-of-flight (dToF) sensors requires training data that captures the nuances of real-world hardware. To this end, our simulator implements a high-fidelity model of an Active Single-Photon Camera (ASPC) that goes beyond the noise-free depth abstractions of standard game engines. We simulate the complete acquisition pipeline, mapping scene irradiance to noisy photon timestamps, while also modeling on-sensor compression schemes critical for resource-constrained applications like autonomous robots and AR/VR headsets. We discuss the forward imaging model in Section~\ref{subsec:spad}. In this section, explain it in detail along with the implementation details.

\subsubsection{Forward model implementation details}
    
The forward imaging model can be divided into three main stages: (i) computing the photon-transient, which models the geometric and radiometric scene properties, (ii) computing the photon arrival rates, which models the active and ambient light sources, and (iii) computing the LiDAR measurements, which models pileup distortion, compressed readouts, the effect of TDC resolution, and synchronous or free-running mode of operation of SPAD sensors.

\noindent\textbf{Modeling the photon-transient}

The photon-transient $T$ is the scene response function (SRF) modeling the interaction of an ideal laser pulse with the scene. In \lib{} we utilize ground-truth data (RGB albedo and depth maps) generated by the rendering engine. To bridge the discrete spatial domain of the renderer and the temporal domain of the SPAD, we employ a per-pixel Field-of-View (FOV) integration method. For every sensor pixel $(u, v)$, we accumulate the weighted contribution of scene points within its projected cone. The simulator defines a \textit{weighted per-pixel FOV mask}, $\mathcal{M}_{uv}$, which encapsulates the geometric relationship between the scene and the sensor optics\footnote{We explain the forward imaging model for a single pixel and drop the subscript $uv$.}. For each pixel, the discrete \emph{scene response} $T[b]$ at bin index $b$ aggregates the radiometric contributions of scene points within a field-of-view mask $\mathcal{M}$ as
\begin{equation}
    T[b] = \eta_\text{opt} \sum_{p \in \mathcal{M}} \mathbb{I}\big(b \Delta t \le t_p < (b+1)\Delta t\big)\, \frac{\alpha_p}{z_p^2}\, W(p),
    \label{eq:transient_supp}
\end{equation}
where $\Delta t$ is the temporal bin width, $\alpha_p$ is albedo, $1/z_p^2$ the inverse-square falloff, $W(p)$ the mask weight, and $\eta_\text{opt}$ the quantum efficiency. This mask-based approach allows us to simulate realistic lens effects, such as vignetting and blooming, which are critical artifacts in robotic perception pipelines.

\noindent\textbf{Modeling the active and ambient light sources}

The raw scene transient assumes an ideal impulse illumination. To model the active source, we convolve the transient with the system's Instrument Response Function (IRF) represented by $g (t)$. The simulator supports parametric pulse shapes (e.g., Gaussian, Square) or empirical IRFs derived from hardware calibration. Robotic platforms often operate in uncontrolled lighting (e.g., direct sunlight). We model ambient light not as additive noise, but as a distinct photon source. The simulator configures \emph{spectral bandpass filters}, ensuring that only ambient light overlapping with the laser's spectrum contributes to the background photon count rate, modeled as a DC offset $\lambda_\text{ambient}$. The resulting computation represents the expected photon arrival rate $\lambda(t)$ incident on the detector:

\begin{equation}
    \lambda(t) = (T * g)(t) + \lambda_\text{ambient}
\end{equation}

\noindent\textbf{Modeling sensor properties and compressed readouts}

\begin{figure*}[t]
  \begin{center}
    \includegraphics[width=1.0\textwidth]{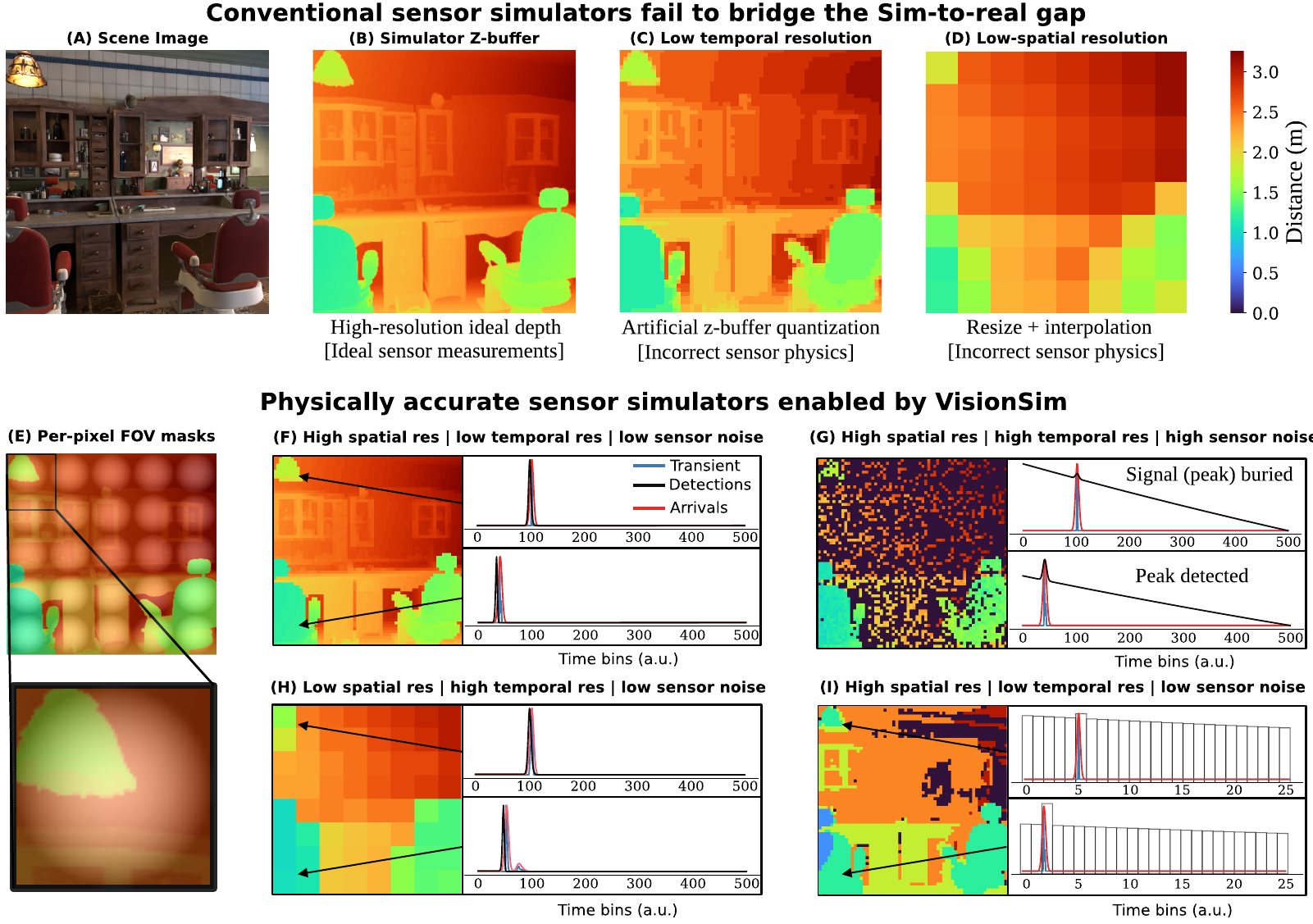}
  \end{center}
  \vspace{-1em}
    \caption{\textbf{\lib{} enables physically accurate 3D sensor simulations:} Conventional simulators create idealistic depth maps (\textbf{B}) from scene data (\textbf{A}). They fail to bridge the sim-to-real gap, ignoring sensor physics by emulating low temporal resolution (\textbf{C}) and low spatial resolution (\textbf{D}) via naive quantization. \lib{} enables physically accurate emulation of sensor properties such as per-pixel FOVs (\textbf{E}); high spatial resolution with low sensor noise (\textbf{F}); dead time effects (\textbf{G}) and hardware constraints, including low spatial (\textbf{H}) and temporal (\textbf{I}) resolution.}
  \label{fig:active_spc_features}
\end{figure*}

To transform continuous photon rates into discrete digital measurements, \lib{} models the stochastic behavior of sensor electronics along with realistic hardware readout schemes. First, photon arrivals are sampled from an inhomogeneous Poisson process, where the detected count $k$ within a given time bin is distributed as $k \sim \mathcal{P}(\lambda(t))$ to account for both active signal and ambient background flux. We explicitly incorporate a non-paralyzable dead time ($\tau_{dead}$) model to simulate the non-linear pileup distortions typically observed when operating in high-flux scenarios. We support two different on-sensor compression methods: (i) standard equi-width histogramming (EWH), which aggregates arrivals into fixed temporal intervals, used in conventional LiDARs, and (ii) equi-depth histogramming (EDH). Designed to mitigate bandwidth constraints in high-resolution arrays, EDH utilizes adaptive quantization to dynamically vary bin widths based on local photon density, preserving high temporal resolution for sharp, peaky distributions while compressing sparse background noise. Suppl.~Fig.~\ref{fig:active_spc_features} shows the effect of different readout schemes, varying FOV mask, sensor temporal resolution and sensor dead time. 

\noindent\textbf{PyTorch implementation with unit-aware configuration system}
\label{subsubsec:implementation}

The simulator is implemented entirely in PyTorch, prioritizing modularity and vectorized operations. By representing the sensor measurements as a high-dimensional tensor, we leverage massive parallel operations. Unlike CPU-bound simulators that iterate over pixels, our vectorized implementation processes mega-pixel arrays simultaneously on the GPU, significantly reducing the simulation time for dense SPAD arrays. 

Unlike rigid legacy simulators, our framework utilizes a strict, unit-aware configuration system (powered by \texttt{pint} and \texttt{ruamel.yaml}). This ensures that all simulation parameters, from laser power (Watts) to pixel pitch (Micrometers), are dimensionally consistent, eliminating physical unit errors before runtime. 

Another distinguishing feature of our framework is the fully differentiable execution modes, which use PyTorch's autograd functionality. This formulation allows gradients to flow from the downstream task loss (e.g., depth reconstruction error) back to the sensor parameters. This enables end-to-end optimization of hardware parameters, such as the sensor IRF or dead time.

\subsubsection{Physically Constrained Single Photon LiDAR Simulations using \lib{}}
Our goal with \lib{} is to bridge the sim-to-real gap, and to do so, we need to be able to model sensor physics realistically. The reason conventional simulators fail to do this is that they create ideal, high-resolution depth maps from scene data. They fail to model limitations in sensor performance as we show in Fig. \ref{fig:active_spc_features} (\textbf{A}, \textbf{B}, \textbf{C}, \textbf{D}). The tradeoff between bandwidth and resolution is something every single-photon LiDAR designer needs to consider when developing a system. Lowering spatiotemporal resolution offers a balance between the data bottleneck and fidelity. However, to evaluate the design choices, conventional simulators use techniques such as artificial quantization and resize-interpolation, which do not adequately
model the physical world. \lib{} enables physically accurate emulation of realistic sensor and optical properties. Complex optical effects can be modeled using per-pixel FOV masks, shown in Fig. \ref{fig:active_spc_features} (\textbf{E}), including vignetting, a process largely ignored by conventional simulators. The FOV masks also allow us to emulate sensors of varying spatiotemporal resolution while retaining a physically accurate acquisition model (Fig. \ref{fig:active_spc_features} \textbf{F}, \textbf{H}).

\lib{} includes environment non-idealities during acquisition, such as noise from ambient light (Fig. \ref{fig:active_spc_features} \textbf{G}) as well as non-idealities unique to this modality. One example of this stems from the implications of dead time. This is a requirement of these sensors because their ability to detect single photons stems from their high sensitivity to energy introduced to the system. They require time to return to an equilibrium state before additional detections. An unavoidable problem arises from this physical process, which is known as pileup. Pileup occurs when photons are not detected during a SPAD's dead time, which leads to two problems. Firstly, photon counts are underestimated. Secondly, these early detections bias the sensor towards earlier timing bins and can cause a reported depth to be lower than an object's true depth (referred to as range-walk error) as demonstrated in Suppl.~Fig.~\ref{fig:active_spc_features}.

\subsection{Indirect Time-of-Flight Simulation}
Unlike direct time-of-flight (dToF) systems that measure the timestamps of individual photon detections, indirect time-of-flight (iToF) cameras estimate depth by emitting amplitude-modulated light and analyzing the phase shift of the reflected signal relative to the source~\cite{lange2001}. An iToF system consists of a temporally modulated light source with radiant intensity $M(t)$ and a sensor with a temporally modulated exposure (demodulation) function $D(t)$. Let $\Gamma(p)$ be the scene depth at pixel $p$. The radiance $L(p, t)$ incident at pixel $p$ at time $t$ is given by~\cite{GuptaToG_2018}:
\begin{equation}
    L(p, t) = \beta(p) M\left(t - \frac{2\Gamma(p)}{c}\right) + L_a(p),
\end{equation}
where $c$ is the speed of light, $\beta(p)$ encapsulates the scene albedo, reflectance, and inverse-square fall-off, and $L_a(p)$ is the ambient light component. 

The brightness $B(p)$ at pixel $p$ over an integration time $\tau$ is the correlation between the incoming radiance and the demodulation function:
\begin{equation}
    B(p) = \int_0^\tau D(t) L(p, t) dt = \beta(p) F(\Gamma) + A(p),
\end{equation}
where $A(p) = L_a(p) \int_0^\tau D(t) dt$ is the ambient component. $F(\Gamma)$ is the normalized correlation function between the modulation and demodulation functions:
\begin{equation}
    F(\Gamma) = \frac{1}{M_{total}} \int_0^\tau D(t) M\left(t - \frac{2\Gamma(p)}{c}\right) dt,
\end{equation}
where $M_{total} = \int_0^\tau M(t) dt$ is the total flux emitted by the source. Finally, the measured brightness $\tilde{B}(p)$ is corrupted by shot noise, quantization and other non-idealities as governed by the sensing technology.

By varying the modulation and demodulation functions across multiple phase offsets, the sensor captures several correlation measurements, commonly referred to as taps, to reconstruct the depth $\Gamma(p)$. The top row of Fig.~\ref{fig:itof} illustrates this acquisition process under ideal conditions. Here, the emulated iToF camera is using a conventional 3-tap delta-sinusoidal modulation/demodulation scheme (others such as square, ramp, Hamiltonian, and Hilbert~\cite{GuptaToG_2018} are also available within \lib{}) with a modulation frequency of $20$ MHz, showing the ground truth RGB scene, the corresponding ground truth depth map, and the ideal correlation measurements across three phase taps.

However, real-world iToF systems are highly susceptible to physical non-idealities and environmental factors that conventional computer vision simulators fail to model. \lib{} addresses this gap by implementing physically accurate emulation of these effects, enabling the validation of downstream depth estimation algorithms under realistic conditions. As shown in the bottom row of Fig.~\ref{fig:itof}, we simulate several critical iToF failure cases and demonstrate their direct impact on depth reconstruction. All examples in the figure are the result of varying a single simulation parameter, and these effects can of course be combined:

\begin{myitemize}
    \item \textbf{Incorrect Modulation Frequency:} The modulation frequency defines the unambiguous depth range of the sensor. In the figure, the incorrect modulation frequency uses $35$ MHz (yielding an unambiguous depth of only $\tildeNice$$4.3$ meters, which is not enough for the scene). The phase wraps around, causing severe phase wrapping ambiguities where distant objects appear deceptively close.
    \item \textbf{Sensor Saturation:} When the exposure time is too long, or an object is highly reflective or positioned too close to the active light source, the accumulated charge can exceed the sensor's full well capacity. This saturates the pixel and corrupts the correlation measurements. Here, clipping occurs because the full well capacity of $50$k electrons is exceeded. This injects non-linear distortions that corrupt the reconstructed depth (notice the front row chairs and ceiling lamps).
    \item \textbf{Ambient Light Corruption:} Strong background illumination introduces a significant DC offset to all taps. In this example, the ambient light level is raised to match the power of the active source. This degrades the modulation contrast and signal-to-noise ratio, leading to incorrect depth reconstruction, particularly for distant objects as they are more dimly illuminated by the active source.
    \item \textbf{Motion Blur:} Because iToF systems must sequentially capture multiple frames (taps) to reconstruct a single depth map, any relative motion between the camera and the scene during the acquisition sequence leads to multi-tap motion artifacts. For this example, motion blur happens over $100$ms. \lib{} simulates this effect by temporally offsetting the frames used for each tap, accurately reproducing the resulting depth tearing and ghosting at object boundaries.
    \item \textbf{Sensor Noise and Quantization:} We model physical noise sources such as shot noise and read noise, alongside the finite resolution of the analog-to-digital converters (quantization). In the figure, the noise corresponds to shot noise plus a read noise of $10e^-$ and a $12$-bit ADC. These factors lead to high-frequency depth noise and quantization banding in the reconstructed depth map.
\end{myitemize}
By modeling these physical degradation processes, \lib{} provides a reliable playground for developing and stress-testing robust phase-unwrapping, de-noising, and high-dynamic-range iToF depth reconstruction pipelines.

\subsection{Thermal Cameras}
\label{sec:thermal-rendering}
Long-wave infrared (LWIR, $8$--$14\,\mu\mathrm{m}$) thermal cameras occupy a fundamentally different place in the imaging stack than the other modalities in VisionSIM: rather than measuring light scattered off the scene within the last microseconds, they measure the long-wave radiation a surface \emph{emits} as a consequence of its temperature. The world simulation layer supplies that temperature field through the integrated heat-transport solver (Sec.~\ref{sec:thermal-heat-transport}); here we describe how the sensor emulation layer turns it into a measurement in two stages: rendering the temperature field into in-band thermal radiance with a graybody shader (Sec.~\ref{sec:thermal-graybody}), and applying the microbolometer image-formation model whose thermal inertia gives real LWIR video its characteristic lag and smear (Sec.~\ref{sec:thermal-microbolometer}).

\subsubsection{Rendering the Temperature Field into Thermal Radiance}
\label{sec:thermal-graybody}
The solver produces a per-vertex temperature field, which is not yet an image. To turn it into the in-band radiance a thermal camera would collect, we render the scene with a physically motivated \emph{graybody} emission model. Rather than maintaining a separate thermal copy of the scene, we render thermal imagery from the \emph{same} scene, camera, and animation as the RGB render by toggling a small set of scene-level overrides---swapping each material for its thermal counterpart, and by disabling visible-band lamps which an LWIR camera does not see. This guarantees that the RGB and thermal images are pixel-aligned and that any edit to the scene propagates to both modalities.

\paragraph{Graybody emission shader.}
At LWIR wavelengths each surface radiates as an opaque graybody, contributing both its own thermal emission within the camera's sensitivity band and a reflection of the long-wave radiation arriving from the rest of the scene. We model the outgoing in-band radiance at $\mathbf{x}$ as
\begin{equation}
    L_\mathrm{out}(\mathbf{x}) \;=\; \epsilon(\mathbf{x})\,U\!\big(T(\mathbf{x})\big) \;+\; \big(1-\epsilon(\mathbf{x})\big)\,L_\mathrm{in}(\mathbf{x}),
    \label{eq:thermal-shader}
\end{equation}
where $T(\mathbf{x})$ is the surface temperature from the heat solver, $\epsilon(\mathbf{x})\in[0,1]$ is the LWIR emissivity, $L_\mathrm{in}(\mathbf{x})$ is the incident long-wave radiance gathered from the rest of the scene~\cite{Narayanan_2026_CVPR}, and $U(T)$ is the camera response to a blackbody at temperature $T$, obtained by integrating Planck's law against the camera's spectral response and generally well-approximated by the Sakuma--Hattori model~\cite{sakuma1982establishing, Narayanan_2026_CVPR}. The first term is the graybody emission within the LWIR specturm, scaled by emissivity through Kirchhoff's law; the second is its conjugate reflection, weighted by $1-\epsilon$, which is exactly the reflectivity an opaque graybody assigns by energy conservation. Although the angular reflectance of real materials in the $8$--$14\,\mu\mathrm{m}$ band is not as well characterized as in the visible, we adopt a Lambertian BRDF for the reflected term: it is the canonical choice for a graybody, keeps the model fully determined by the single per-surface parameter $\epsilon$, and is straightforwardly replaceable by a measured wavelength-specific BRDF as the literature matures. The two terms together capture the qualitative behavior of real LWIR imagery without any material-specific special cases: a polished low-$\epsilon$ metal ``looks cold'' because its mix weights the reflection of its cooler surroundings, while a high-$\epsilon$ matte surface reads its own emitted light almost directly.

\subsubsection{Microbolometer Thermal Inertia}
\label{sec:thermal-microbolometer}
A real uncooled LWIR camera does not record the instantaneous in-band radiance produced by the renderer. The dominant commercial LWIR detector is the \emph{microbolometer}: an array of thermally isolated absorber membranes, one per pixel, whose temperature is set by the incident radiation and read out as an electrical resistance. Because the membrane has finite heat capacity, its temperature cannot change instantly with the incident radiation; the readout instead lags behind, and that lag is what gives real thermal video its characteristic motion smear and ghost trails behind moving hot objects~\cite{Srinivasan_Ramanagopal_2020}.

Quantitatively, a single absorber membrane is a lumped thermal mass connected to a substrate at temperature $T_s$ through legs of finite thermal conductance, obeying the first-order ODE \cite{Srinivasan_Ramanagopal_2020,vollmer2020infrared,topaloglu2009analysis}
\begin{equation}
    \frac{C_\mathrm{th}}{G_\mathrm{th}}\,\frac{\mathrm{d}\,\Delta T(t)}{\mathrm{d}t} \;+\; \Delta T(t) \;=\; \frac{\alpha_\mathrm{abs}}{G_\mathrm{th}}\,\phi(t),
    \label{eq:bolometer-ode}
\end{equation}
where $\Delta T(t)=T_m(t)-T_s$ is the membrane-to-substrate temperature difference, $C_\mathrm{th}$ is the membrane heat capacity ($\mathrm{J}/\mathrm{K}$), $G_\mathrm{th}$ is the thermal conductance of its supports ($\mathrm{W}/\mathrm{K}$), $\alpha_\mathrm{abs}$ is the membrane's in-band absorptivity, and $\phi(t)$ is the instantaneous incident in-band radiative power ($\mathrm{W}$). The ratio $\tau = C_\mathrm{th}/G_\mathrm{th}$ has units of time and is the membrane's \emph{thermal time constant}, the natural timescale on which it equilibrates to a change in incident radiation; for commercial uncooled LWIR detectors, $\tau$ typically lies in the $5$--$15\,\mathrm{ms}$ range. Defining the hypothetical instantaneous steady state $T_m^\mathrm{ss}(t) = T_s + (\alpha_\mathrm{abs}/G_\mathrm{th})\,\phi(t)$---the temperature the membrane \emph{would} reach if it equilibrated infinitely fast to the radiation present at time $t$---Eq.~\ref{eq:bolometer-ode} takes the canonical first-order form
\begin{equation}
    \tau\,\frac{\mathrm{d}\,T_m(t)}{\mathrm{d}t} + T_m(t) \;=\; T_m^\mathrm{ss}(t).
    \label{eq:bolometer-tau}
\end{equation}
The interpretation is direct: the membrane temperature $T_m(t)$ chases its instantaneous target $T_m^\mathrm{ss}(t)$ at a rate set by $\tau$. The readout integrated circuit then reports a digital count $I(t)=P_1\,T_m(t)+P_2$ that is affine in $T_m$, so up to a one-time factory calibration the reported video is a temporally low-pass-filtered version of the scene radiance. Crucially, $T_m^\mathrm{ss}(t)$ is exactly the idealized infinite-bandwidth measurement that a perfect (zero-$\tau$) camera would produce---which is precisely the in-band radiance $L_\mathrm{out}$ that our graybody renderer writes into each frame (Eq.~\ref{eq:thermal-shader}). We can therefore synthesize a realistic microbolometer response by filtering the rendered radiance sequence through Eq.~\ref{eq:bolometer-tau}.

\paragraph{Closed-form solution and discretization.}
Equation~\ref{eq:bolometer-tau} is a linear first-order ODE; multiplying by the integrating factor $e^{t/\tau}$ and integrating yields the convolution of the ideal radiance against a one-sided exponential kernel~\cite{Srinivasan_Ramanagopal_2020},
\begin{equation}
    T_m(t) \;=\; \frac{1}{\tau}\int_{-\infty}^{t} e^{-(t-s)/\tau}\,T_m^\mathrm{ss}(s)\,\mathrm{d}s.
    \label{eq:bolometer-conv}
\end{equation}
Equation~\ref{eq:bolometer-conv} says that the current membrane temperature is a weighted average of all past radiance, with weights given by a decaying exponential of timescale $\tau$: recent radiance contributes strongly, while radiance from many $\tau$ ago has been ``forgotten.'' This is the precise sense in which the bolometer acts as a temporal low-pass filter on the scene.

To apply this to rendered video, we discretize Eq.~\ref{eq:bolometer-conv} at the frame rate. Let $t_n = n\,\Delta t$ with $\Delta t = 1/\mathrm{fps}$, let $x[n] := T_m^\mathrm{ss}(t_n)$ denote the rendered input at frame $n$, and let $y[n] := T_m(t_n)$ denote the filtered output we want to compute. Splitting the convolution integral at the previous frame time $t_{n-1}$ and factoring the exponential kernel as $e^{-(t_n-s)/\tau} = e^{-\Delta t/\tau}\,e^{-(t_{n-1}-s)/\tau}$ gives
\begin{equation}
    y[n] \;=\; e^{-\Delta t/\tau}\,y[n-1] \;+\; \frac{1}{\tau}\int_{t_{n-1}}^{t_n} e^{-(t_n - s)/\tau}\,T_m^\mathrm{ss}(s)\,\mathrm{d}s,
    \label{eq:bolometer-split}
\end{equation}
where the first term is exactly what the convolution from $-\infty$ to $t_{n-1}$ contributes, attenuated by one frame's worth of exponential decay. The renderer samples the scene once per frame with no within-frame motion blur, so we model the input as piecewise-constant across each interval (a zero-order hold), $T_m^\mathrm{ss}(s) \approx x[n]$ for $s\in(t_{n-1}, t_n]$. Pulling $x[n]$ out of the integral and evaluating $\frac{1}{\tau}\int_0^{\Delta t} e^{-u/\tau}\,\mathrm{d}u = 1 - e^{-\Delta t/\tau}$ collapses Eq.~\ref{eq:bolometer-split} to a single-pole infinite impulse response low-pass filter,
\begin{equation}
    y[n] \;=\; a\,y[n-1] + (1-a)\,x[n], \qquad a = e^{-\Delta t/\tau}.
    \label{eq:bolometer-iir}
\end{equation}
For piecewise-constant input this is the \emph{exact} integrator of Eq.~\ref{eq:bolometer-tau}, not a finite-difference approximation, and the only free parameter is the single dimensionless ratio $\Delta t/\tau$ that determines $a$. The limits are sanity-checking: $\tau\to 0$ (instant camera) gives $a\to 0$ and $y[n] = x[n]$, so the output equals the input with no lag; $\tau\to\infty$ (infinitely sluggish camera) gives $a\to 1$ and $y[n]=y[n-1]$, so the output is frozen. In the regime of interest, at $60\,\mathrm{fps}$ and $\tau = 10\,\mathrm{ms}$, $a\approx 0.19$ and a hot pixel reaches $\sim\!90\%$ of its steady response in about three frames; larger $\tau$ stretches the lag proportionally. When $\tau \ll \Delta t$ (a fast detector under a slow render) the zero-order hold becomes inaccurate, and we instead linearly interpolate the input between rendered frames and run $S$ substeps per frame with $a' = e^{-\Delta t/(S\tau)}$, recovering Eq.~\ref{eq:bolometer-iir} at $S=1$.

\paragraph{Scope of the thermal sensor model.}
Our graybody shader (Eq.~\ref{eq:thermal-shader}) assumes a Lambertian BRDF and a single scalar emissivity, whereas real materials exhibit angularly and spectrally varying reflectance in the $8$--$14\,\mu\mathrm{m}$ band that is far less characterized than in the visible. On the detector side, our current model captures only the microbolometer's thermal inertia, the first-order temporal low-pass that produces motion smear, and not the temporal noise families that define the NETD~\cite{NIKLAUS2008168}, the spatial fixed-pattern and column-stripe structure, the focal-plane-array drift between flat-field corrections, vignetting, or narcissus-like self-emission; some of these could be layered onto the rendered radiance as a post-process following the standard sensor signal chain, which we leave to future work.

\section{Building Digital Twin of Commercial dToF LiDAR}
\label{app:digital_Twin}
To demonstrate \lib{}'s ability to bridge the real-to-sim gap we conducted an experiment in which we implemented a \textit{perceptual} digital twin and verified its accuracy with respect it's real world counterpart (as described in Sec.~\ref{sec:applications}). 

\subsection{Capture Setup}
\label{app:capture}
For this evaluation we captured the real scene using an off-the-shelf, commercial, single photon LiDAR developed by ADAPs Photonics \cite{adaps_ads6311} (the ADS6311) used because it gives access to raw unprocessed histograms. The ADS6311 is a 1D scanning LiDAR with a resolution of $256\times192$ with  $3\times3$ spatial binning to improve SNR (i.e. there are $768\times576$ SPAD elements). This device has a dead-time of $\tildeNice 11$ns and a temporal resolution of $754$ps. The transmitter uses VCSEL laser diodes to scan a 6-row-wide band in the vertical direction. 

We set up a scene in lab with a retroreflective stop sign. Retroreflector's send incident light back towards its source resulting in two phenomena: pileup and glare (referred to as bloom in the LiDAR community). Pileup is the result of a SPAD element's "dead-time" which biases the sensor toward earlier detections. In the event of strong returns such as those from a retroreflector, this bias is strengthened which leads to what is referred as range walk error (an error in the measured depth) as well as a reduction in measured intensity (due to photons in later time bins being missed). In addition to causing pileup, some of the returning waveform scatters diffusely on the lens as well as bouncing between the lens, receiver housing, and pixel array which all lead to spurious detections. This is the cause of what is known as glare or bloom (in traditional CMOS imaging this is sometimes called lens flare). 

This evaluation gave us the ability to determine if \lib{} could faithfully reconstruct the artifacts stemming from the physical acquisition process which is demonstrated in Figure ~\ref{fig:applications}. 

\subsection{Blender Simulated Scene}
\label{app:real2sim}
We then reconstructed this scene in Blender with the stop sign and background at roughly the same depth and dimensions. Also, with the camera at the same position as it was when we captured the real scene. This is not a perfect reconstruction as the ADAPs kit provides only opencv-style coefficients which Blender does not follow for their barrel distorted lenses. Thus, we used a lens and camera model that was a perceptually similar approximation to the ADS6311. Additionally, currently \lib{} does not directly support retroreflective objects; to overcome this we used the material identifiers and manually set albedo to approximate the extremely high albedo of retroreflectors (as well as the comparatively low albedo of the background, a black canvas tarp).   

\subsection{Learning the Impulse Response Function}
\label{app:parameter_learning}
To evaluate \lib{}s fully differentiable pipeline, we started with an initial assumption of an impulse response function that is a perfect delta (with some small positive constant in the remaining bins to avoid zero gradients). For learning the IRF we used a single scene that just contained the background tarp. In practice, a more controlled setup with known albedo targets should yield even better results. We compared a simulated histogram to the true measured histograms using a cross-correlation loss function (over 1000 epochs) given by:
\begin{equation}
    \label{appEq:correlation_loss}
    Cor_{loss} = \text{mean}({1-\max{(pred * gt)}})  + \lambda TV(b_i - b_{i+1})
\end{equation}
where $pred*gt$ is the pixel-wise cross-correlation between the predicted (simulated) histogram and measured histogram across a $20\times20$ patch, and the total variation term is a smoothness penalty between adjacent bins to avoid sharp spikes in the IRF. The comparison given in Figure~\ref {fig:applications} shows the measured histogram as compared to a simulated histogram using our learned IRF for a point on the stop sign and a point in the region containing glare artifacts.

\begin{figure}
    \centering
    \includegraphics[width=\linewidth]{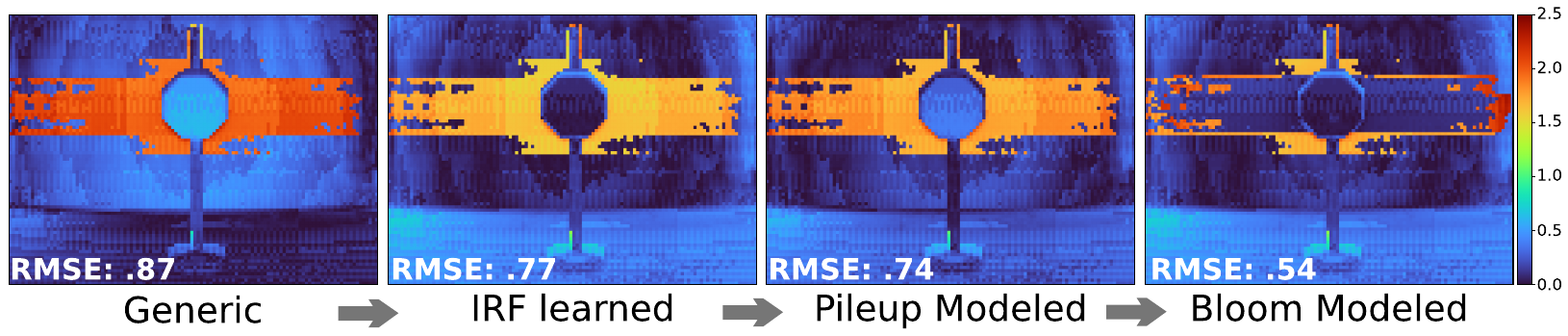}
    \caption{Error Maps for digital twin evaluation; inset: RMSE between the measured depth map and the simulated depth map.}
    \label{fig:error_maps}
\end{figure}
\subsection{Comparison to Real Capture Details}
\label{app:comparison_details}

To simulate the glare artifacts we measured the PSF function of the ADS6311 by disabling the transmitter and pointing an IR flashlight at the same wavelength toward the receiver. We used a small aperture and baffle to ensure light was incident on a single pixel and that there were no multipath effects being measured respectively. These 49 measured PSF's are treated as a basis, with remaining positions interpolated by a distance-weighted sum of the 49 measured positions. We plan to add more robust lens modeling that includes a parametric form of this glare model and can be learned in a similar fashion to the IRF learning described in section ~\ref{app:parameter_learning}. Error maps for each additional modeled effect (IRF, pileup, and bloom) can be found in Figure~\ref {fig:error_maps}. It should be noted that the discrepancy between the reported RMSE from the main paper and this figure is due to a tighter crop around the region of interest. This is in part to highlight the effects of our realistic modeling but also because the background tarp was the most challenging to match in Blender, causing error unrelated to \lib{}s physical model but instead due to imperfections in how the ground truth depth was calculated for the tarp in Blender.

\section{Dataset generation interface}
\label{app:dataset}

\begin{figure}[t]
  \begin{center}
    \includegraphics[width=1.0\textwidth]{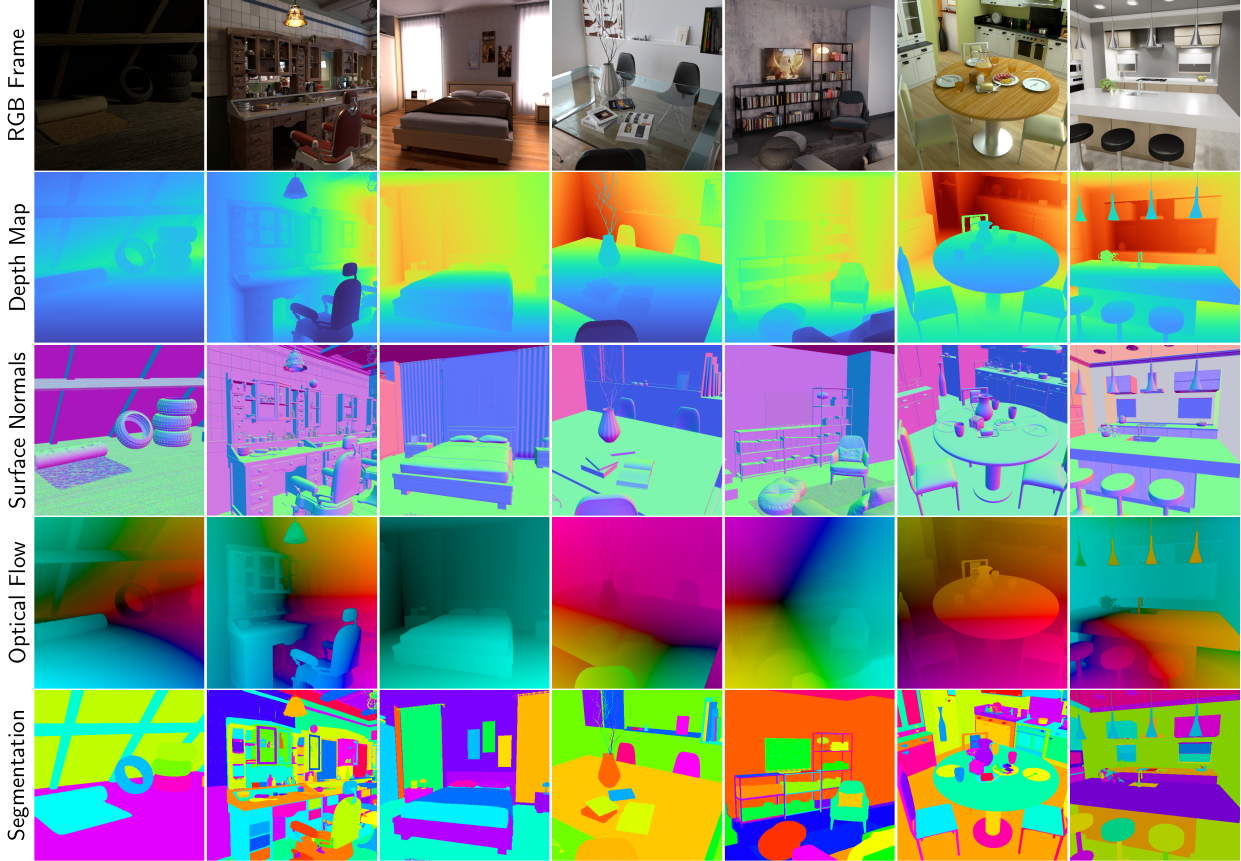}
  \end{center}
  \vspace{-1em}
  \caption{\textbf{Example scenes from the \lib{} dataset.} A collection of diverse indoor scenes featuring different lighting conditions, materials, and ground truth annotations.}
  \label{fig:dataset}
\end{figure}

\noindent\textbf{Asset curation.}
To help accelerate research, we release all $3$D assets and scripts used to create sensor-realistic datasets, lowering the need for manual asset curation, which can be a significant hurdle. Specifically, we release the $100$ high-quality rigged scenes, some of which are shown in Fig.~\ref{fig:dataset}, which have been used to create the \lib{} dataset. These scenes have mostly been crowdsourced from sites such as BlendSwap~\cite{Blendswap} or Blender's example scenes~\cite{blenderexamples}, although some have been made or heavily modified specifically for this use case. All scenes are fully rigged, physics-baked, and validated to render without modification using the Blender version pinned to the corresponding \lib{} release. These assets are released under the Creative Commons license (CC BY 4.0) or similar permissive licenses.

\smallskip
\noindent\textbf{Generation pipeline.}
Dataset generation is driven by a single script, \texttt{scripts/mkdataset.py}, that wraps the same \texttt{RenderConfig} dataclass and \texttt{render\_animation} backend used throughout \lib{}'s CLI. A user specifies a directory of blend-files and an output directory; the script then (i) uniformly samples non-overlapping sub-trajectories from the animation range of each scene, or renders the whole trajectory if desired, (ii) dispatches each \emph{(scene, trajectory)} pair as an asynchronous render job via \texttt{BlenderClients.pool}, and (iii) saves every sequence under a deterministic path which also contains the explicit configuration used to produce the sequence. All relevant metadata are stored in per-modality SQLite databases alongside the raw data. The \texttt{RenderConfig} dataclass exposes every relevant parameter --- resolution, render device, keyframe multiplier, types of ground truth, noise threshold, and per-modality flags --- entirely through the CLI, so a full reconfiguration of the dataset requires no code changes. For example, to render a $800\times800$ dataset at $100$\,fps with depth, normals, optical flow, and segmentation using five parallel jobs:

\begin{verbatim}
python scripts/mkdataset.py create-datasets \
  --scenes-dir=scenes/ --datasets-dir=datasets/ \
  --sequences-per-scene=1 \
  --render-config.width=800 --render-config.height=800 \
  --render-config.depths --render-config.normals \
  --render-config.flows --render-config.segmentations \
  --render-config.keyframe-multiplier=4.0 \
  --render-config.jobs=5
\end{verbatim}

For the full dataset, this script was submitted to HTCondor~\cite{chtc} via a containerized \lib{} image, distributing work across thousands of CPUs and hundreds of GPUs as described in Appendix~\ref{app:world-sim}.  Sensor emulation (interpolation, SPC, event camera, iToF, thermal, IMU) is applied as a post-processing pass on the ground-truth renders using the same \lib{} library functions described in Appendix~\ref{app:sensors}, each of which has a corresponding command-line interface tool.  

\smallskip
\noindent\textbf{Reproducibility.}
Every \lib{} release is containerized with Docker~\cite{merkel2014docker} and tested end-to-end in CI, ensuring that a pinned version of the library and its container can regenerate all data without any environment configuration. Users who wish to reproduce or modify the dataset only need to (i) pull the versioned Docker image corresponding to the release tag, (ii) obtain the released blend-files, and (iii) invoke \texttt{mkdataset.py} with the published configuration.

\end{document}